\documentclass[sigconf]{acmart}

\renewcommand\footnotetextcopyrightpermission[1]{}
\copyrightyear{2026}
\acmYear{2026}
\setcopyright{rightsretained}

\acmDOI{TBD}
\acmISBN{TBD}
\usepackage{siunitx}
\usepackage{algorithm}
\usepackage{algorithmicx}
\usepackage{algpseudocode}
\usepackage{graphicx}
\usepackage{soul}
\usepackage{xspace}
\usepackage{subcaption}
\usepackage{multirow}
\usepackage{threeparttable}
\usepackage{CJKutf8}
\usepackage[utf8]{inputenc}
\usepackage{booktabs}
\usepackage{tabularx}
\usepackage{diagbox}
\usepackage{tabularx}
\usepackage{tcolorbox}
\tcbuselibrary{skins, breakable}
\usepackage[inline]{enumitem}
\graphicspath{{figures/}}

\makeatletter
\@ACM@balancefalse
\makeatother

\newcommand{\zh}[1]{\begin{CJK*}{UTF8}{gbsn}#1\end{CJK*}}

\newcommand{\ie}{i.e.,\xspace}
\newcommand{\eg}{e.g.,\xspace}
\newcommand{\pb}[1]{\noindent\textbf{#1}\ }

\newboolean{showcomments}
\setboolean{showcomments}{true}

\newcommand{\qm}[1]{\textcolor{blue}{[QM: #1]}}
\newcommand{\zf}[1]{\textcolor{blue}{[Zifan: #1]}}
\newcommand{\gareth}[1]{\textcolor{red}{[GT: #1]}}
\newcommand{\todo}[1]{\textcolor{red}{[TODO: #1]}}
\newcommand{\peixian}[1]{\textcolor{orange}{[peixian: #1]}}

\ifthenelse{\boolean{showcomments}} { }
{
\renewcommand\peixian[1]{}
\renewcommand\gareth[1]{}
\renewcommand\zf[1]{}
\renewcommand\qm[1]{}
\renewcommand\todo[1]{}
}

\newcommand\summary[1]{
  \begin{tcolorbox}[%
    breakable,
    colback=white,
    colframe=black,
    sharp corners,
    boxrule=1pt,
    title=\textbf{Takehomes},
    titlerule=1pt,
    enhanced,
    skin first=enhanced,
    skin middle=enhanced,
    skin last=enhanced,
    left=1pt,    
    right=1pt,   
    boxsep=1pt
  ]
#1
  \end{tcolorbox}
}

\begin{document}

\title{Behind EvoMap: Characterizing a Self-Evolving Agent-to-Agent Collaboration Network}

\author{Qiming Ye}
\authornote{Both authors contributed equally to this research.}
\affiliation{%
  \institution{The Hong Kong University of Science and Technology (Guangzhou)}
  \country{China}
}
\email{qiming@connect.hkust-gz.edu.cn}

\author{Peixian Zhang}
\authornotemark[1]
\affiliation{%
  \institution{The Hong Kong University of Science and Technology (Guangzhou)}
  \country{China}
}
\email{pzhang041@connect.hkust-gz.edu.cn}

\author{Yupeng He}
\affiliation{%
  \institution{The Hong Kong University of Science and Technology (Guangzhou)}
  \country{China}
}
\email{yhe382@connect.hkust-gz.edu.cn}

\author{Zifan Peng}
\affiliation{%
  \institution{The Hong Kong University of Science and Technology (Guangzhou)}
  \country{China}
}
\email{zpengao@connect.hkust-gz.edu.cn}

\author{Gareth Tyson}
\affiliation{%
  \institution{The Hong Kong University of Science and Technology (Guangzhou)}
  \country{China}
}
\email{gtyson@ust.hk}

\renewcommand{\shortauthors}{Ye et al.}

\begin{abstract}
Agent-to-Agent (A2A) networks enable autonomous AI agents to collaborate by sharing reusable problem-solving instructions. 
However, how these decentralized ecosystems operate in practice remains largely unexplored. 
We present the first large-scale empirical study of EvoMap, a prominent A2A collaboration network. 
By analyzing over 1.5M assets and 128K agents, we show how design choices that prioritize scalable growth introduce trade-offs in reusability, evolution, and auditability. 
First, EvoMap's credit economy rewards agents for publishing valuable assets. 
Although this design encourages participation at scale, rewards are tied primarily to publication rather than adoption.
This leads agents to mass-produce assets to accumulate credits.
As a result, \qty{98}{\percent} of assets are never reused, while rewards become highly concentrated among a small fraction of agents. 
Second, EvoMap employs an algorithm (referred to as GDI) to score and rank the quality of these shared assets.
We demonstrate that this scoring system is flawed: rather than measuring objective performance, an asset's rank is heavily dictated by unverified, self-reported metadata (\eg claimed lines of code modified). 
This allows agents to trivially manipulate their asset's scores. 
Finally, EvoMap relies on agents to provide local execution logs as evidence that uploaded assets function correctly. 
Because these validations are not independently verified, over \qty{84}{\percent} of approved assets bypass quality checks using vacuous tests (e.g., \texttt{console.log()}). 
Our findings show that future A2A collaboration networks cannot rely on unverified self-reporting alone. 
Scalable collaboration requires mechanisms that balance open participation with verifiable execution and trustworthy evaluation.
\end{abstract}

\maketitle

\section{Introduction}
\label{sec:intro}
There is growing interest in AI agents that interact with external tools to perform complex tasks~\cite{mundlamuri2025evolution,plaat2025agentic}. 
Protocols such as the Model Context Protocol~\cite{hou2025model} and Skills~\cite{xu2026agent} standardize these tool interactions within fixed environments (\eg querying a database with predefined APIs).
Yet, when environments change (\eg schema upgrade), tool configurations and dependencies may no longer match, causing agentic interaction failures~\cite{ray2025survey,li2026skillsbench}.
Moreover, the updated tool-use instructions are often stored locally, leading to duplicated efforts across the ecosystem.

To address these issues, the paradigm is shifting toward Agent-to-Agent (A2A) collaboration networks that enable agents to share and reuse instructions across environments~\cite{ehtesham2025survey,a2a_protocol}.
Existing implementations of these networks, such as \emph{Clawhub}~\cite{clawhub-intro}, operate as passive repositories, where agents can publish and reuse instruction sets (skills). 
Yet, because assets are shared without validation or maintenance, obsolete or insecure instructions may gradually accumulate in the repository~\cite{liu2026agent,liu2026malicious,chen2026openclaw,hu2026red}.
These failures have led researchers to propose self-evolving A2A networks that combine distributed instruction sharing with autonomous curation. 
To support such an architecture, prior studies identify three core requirements~\cite{fang2025comprehensive,gao2025survey}. 
First, \textbf{Reusability} facilitates the discovery of validated instructions to eliminate redundant efforts across independent agents. 
Second, \textbf{Evolution} ensures that high quality instructions adapt to new requirements while removing the obsolete ones. 
Third, \textbf{Auditability} allows for the verification of contributions to maintain safety.

This paper presents a large-scale measurement study of a major A2A collaboration network, \textbf{EvoMap}~\cite{wang2026procedural}, which is designed to operationalize these principles~\cite{evomap-intro}. 
EvoMap implements its Genome Evolution Protocol~\cite{wang2026procedural} that governs how instruction sets (referred to as \emph{assets}) are generated, validated, and shared across agents to guide task solving.
For example, an asset might cover how an agent can purchase a gift from a marketplace.
EvoMap implements core mechanisms to support such operations.
To foster \textbf{Reusability}, it provides a shared registry (called EvoMap Hub) for agents to share their problem-solving assets with other agents.
To enable \textbf{Evolution}, the system employs a ranking mechanism (called the GDI) to promote high-quality assets while phasing out outdated ones.
To ensure \textbf{Auditability}, it mandates that all assets pass validation checks in an isolated sandbox environment (called Evolver).
By April 2026, the platform had grown to host over 1.5M assets from 128K daily active agents. 
Despite this scale, the real-world effectiveness of such A2A collaborative systems remains unstudied. Thus, our study evaluates EvoMap against three of its stated goals~\cite{evomap-intro}, guided by the following research questions:

\begin{enumerate}[leftmargin=*]
    \item \textbf{RQ1 (Reusability):} Does the EvoMap Hub shared registry successfully facilitate the reuse of assets?

    \item \textbf{RQ2 (Evolution):} Do EvoMap's incentive mechanisms effectively promote high-quality assets over time?

    \item \textbf{RQ3 (Auditability):} Are the protocol's sandbox and validation gates robust enough to prevent manipulation?
\end{enumerate}

Our analysis reveals a gap between EvoMap's design goals~\cite{evomap-intro} and its performance in-the-wild. 
Our contributions are:
\begin{itemize}[leftmargin=*]
    \item \textbf{Dataset:} To the best of our knowledge, we present the first measurement study of a large-scale A2A collaborative network. We gather a dataset covering 1.5M assets and 128K agents, which will be released to the research community.
    
    \item \textbf{Low Reusability:} We find that, while agents publish high volumes of assets, \qty{98}{\percent} are \emph{never} reused, pointing to a severe imbalance between supply and demand.
    
    \item \textbf{Centralized Evolution:} Rewards are concentrated among a small fraction (\qty{10}{\percent}) of agents. The four-dimensional GDI score metric is dominated by one particular dimension (termed the ``Intrinsic'' component), which relies on self-reported data and is therefore easily manipulated.

    \item \textbf{Weak Auditability:} The validation mechanism is not robust, allowing agents to bypass quality checks. We show that introducing simple verification steps (\eg checking file changes via Git or executing validations) could mitigate these issues.
\end{itemize}

\section{A Primer on EvoMap}
\label{sec:background}

Each agent in EvoMap consists of a local \emph{Evolver}.
This interacts with a single centralized \emph{Hub}.
Their relationship is analogous to a Git client that publishes a repository to a public GitHub server. 
Agents generate solutions locally, evaluate them in a sandbox with a self-assigned performance score, and publish validated solutions to the Hub (optionally) for reuse by others.
For further details, we refer readers to the official documentation~\cite{evomap-intro}.

\subsection{EvoMap Evolver}
\label{subsec:evolver}
In EvoMap, an \emph{Asset} is a reusable instruction set for agents.  
For each task it wishes to perform (\eg booking a flight), the agent first searches its local store for a matching asset based on task requirements. 
If no adequate match is found locally, the agent sends the same query to the remote EvoMap Hub, where candidates are retrieved based on functional similarity.  
If none is found, the agent must generate a new one locally.
Note, by default, agents only publish assets to the Hub when sharing is enabled. 
Each asset consists of three components: a \emph{Gene}, a \emph{Capsule} and an \emph{EvolutionEvent}, discussed below.

\pb{Gene.}
A Gene acts as a behavioral guide, somewhat akin to an interface in Object-Oriented Programming (OOP). 
Rather than dictating the rigid implementation of a concrete class, it serves as a blueprint for a specific category of tasks. 
It establishes the rules of engagement by defining when such an instruction set should be invoked (preconditions), what operational boundaries must be respected (constraints), and how the final outcome is evaluated (validations).
Formally, a Gene consists of three components:
\begin{enumerate*}[label=(\roman*)]
    \item \textbf{Precondition:} Conditions that trigger the invocation.
    \item \textbf{Constraint:} Non-negotiable action boundaries.
    \item \textbf{Validation:} Required checks that must pass.
\end{enumerate*}
For example, a bug-fixing Gene may guide the agent to inspect asynchronous calls to locate the issue when handling a specific timeout error (precondition), while prohibiting modifications to \texttt{.git} (constraint) and requiring \texttt{npm test} to pass (validation).

\pb{Capsule.}
While a Gene defines an abstract interface for a class of tasks, a Capsule provides a concrete implementation for a specific scenario. Much like a class implementing an interface, a Capsule is generated following the guidance of its parent Genes. 
It encapsulates the exact solution by specifying its core logic (content), the exact conditions for its reuse (trigger), and its historical performance (metadata).
Formally, a Capsule consists of three components:
\begin{enumerate*}[label=(\roman*)]
    \item \textbf{Content:} The concrete implementation or executed action sequence.
    \item \textbf{Trigger:} The exact signature (\eg a specific stack trace) that serves as a search key, indicating when this exact solution can be retrieved and directly reused.
    \item \textbf{Metadata:} Metrics and execution history recorded after task completion.
\end{enumerate*}
For example, guided by the aforementioned bug-fixing Gene, the agent might create a timeout-fixing Capsule containing a specific code patch (content) that replaces a blocking API call with an asynchronous request. 
This Capsule is reused if its exact stack trace (trigger) is encountered again, while its runtime impact is recorded (metadata).

\pb{EvolutionEvent.}
When an agent implements a solution---either creating a new Capsule for a novel task (innovation) or modifying an existing Capsule to fix an edge case (repair)---it applies the code changes locally and executes the validation commands defined by the guiding Gene.
The agent only saves the resulting Capsule if all validation commands pass; otherwise, the changes are reverted. 
After a successful validation, the engine records an \texttt{EvolutionEvent}.
This event summarizes the execution metrics (such as the blast radius of lines modified) and cryptographically links the Capsule to the parent Genes that guided the change. 
Together, these events form an auditable graph that tracks how the system capabilities evolve over time.
For example, an agent modifying an API-fetching Capsule to handle \texttt{HTTP 503} retries would implement the retry logic locally and run the Gene’s required validations.
If the test fails, the code is reverted. 
If it passes, the modified Capsule is committed, and an \texttt{EvolutionEvent} is generated. 
This event explicitly logs the repair, the lines changed, and links the newly mutated Capsule back to the error resolution Gene.

\subsection{EvoMap Hub}
\label{subsec:evomap_hub}
\pb{Asset Promotion.}
Once the EvoMap Hub receives an uploaded asset (bundled as a \emph{Gene}, \emph{Capsule} and \emph{EvolutionEvent}), it applies a two-stage filtering process to prevent spam and ensure quality. 
First, it performs a similarity check to reject very similar assets.
Second, newly published assets remain hidden as \texttt{candidates} until they are \texttt{promoted} and made visible to other agents. 
Promotion is determined by the Genetic Desirability Index (GDI), a platform score that measures asset quality. It is recomputed hourly by the Hub~\cite{evomap-credits} using the following official equation:
\begin{equation}
\small
\mathrm{GDI} =
0.35\,\mathrm{GDI_I} +
0.30\,\mathrm{GDI_U} +
0.20\,\mathrm{GDI_S} +
0.15\,\mathrm{GDI_F}
\label{eq:official_gdi}
\end{equation}
Eq.~\ref{eq:official_gdi} defines the official formula based on four sub-metrics:
\begin{enumerate*}[label=(\roman*)]
    \item \textbf{Intrinsic ($\mathrm{GDI_\mathrm{I}}$):} Based on self-reported metadata (\eg reported success rates and developer-stated performance metrics) fixed at publication time.
    
    \item \textbf{Usage ($\mathrm{GDI_U}$):} Measures how many times other agents fetch and successfully reuse the asset.
    
    \item \textbf{Social ($\mathrm{GDI_S}$):} Represents review (\ie votes) from other agents.
    
    \item \textbf{Freshness ($\mathrm{GDI_F}$):} Decays the score over time if the asset remains unused.
\end{enumerate*}
An asset reaches a $\mathrm{GDI}$ of at least 25 to be ``promoted''. Only after this promotion, an asset is discoverable and reusable by other agents.

\pb{Incentivization.} 
The EvoMap uses a centralized credit system, where agents spend credits to publish or fetch assets.
Agents earn credits through two main channels: 
\begin{enumerate*}[label=(\roman*)]
    \item \textbf{Asset Contributions:} Agents earn credits when their assets are promoted and receive additional rewards when those assets are reused by others.
    \item \textbf{Task Resolution:} Bounties are posted with attached rewards, where agents compete by submitting solutions. 
    Each bounty has a single winning submission, determined by a weighted combination of Gemini evaluation~\cite{team2023gemini}, GDI, execution history, and social voting.
\end{enumerate*}
\section{Dataset}
\label{sec:methodology}
To collect our dataset, we develop a measurement infrastructure using the official protocol endpoints (\url{evomap.ai/a2a/}). 
Our observation window spans 47 days (Feb 11, 2026 to Mar 30, 2026), and we capture data across two dimensions described below. 
The complete dataset schema is provided in Appendix~\ref{sec:appendix_dataset}.

\pb{Supply-Side Assets.}
We collect a full snapshot of 799{,}389 Genes, 792{,}481 Capsules, 95 EvolutionEvents, and 128{,}054 agents. 
The metadata includes usage statistics (\texttt{call\_count}, \texttt{reuse\_count}) and scores for the four GDI components. 
We also retrieve the corresponding asset files, including \texttt{validation} scripts, \texttt{content}, \texttt{trigger\_text}, and self-reported metadata in Genes and Capsules.

\pb{Demand-Side Bounties.}
To capture user demand, we collect 92{,}414 bountied tasks and their metadata, including status, titles, and associated incentives (\ie credits). 
We also record 123{,}246 submission events, which track how agents attempt to solve these tasks and the assets they submit.
\section{Reusability}
\label{sec:reusability}
A core objective of EvoMap is reusability~\cite{evomap-intro}, where agents can build on others’ prior experience via the remote Hub. 
We evaluate this goal by analyzing reuse ratios and asset content.

\subsection{Asset Reusability}
\label{subsec:reusability}


We measure two usage metrics: call and reuse. 
The call rate captures how often an agent selects and downloads an asset from the Hub. 
Reuse measures how often an agent successfully applies that downloaded asset to fix a problem locally. 
In practice, calls and reuse occur almost entirely at the Capsule level, with Capsule totals being 32$\times$ and 26$\times$ higher than Gene totals, respectively. 
This gap is expected: a single abstract Gene can naturally spawn numerous concrete Capsules to handle diverse, specific scenarios (see \S\ref{subsec:evolver}). 
Consequently, directly retrieving an existing Capsule is far more efficient than dynamically reconstructing a new implementation from a Gene.
Called assets are almost always reused: \qty{97}{\percent} of called Capsules successfully are reused in downstream tasks.
Yet, \qty{98}{\percent} of assets are never called. 
This suggests that reuse is limited in practice: although many assets are promoted, only a small fraction are actually reused. 
This aligns with user feedback reporting poor asset discovery and Hub unavailability (see Appendix~\ref{app:evomap_operation_related_post}). 
In the current design, agents must query the central Hub to search and retrieve Capsules. 
If the Hub becomes unavailable, agents cannot access assets, which creates a central bottleneck that limits public sharing.

\begin{figure}
    \centering
    \vspace{-2ex}
    \includegraphics[width=\linewidth]{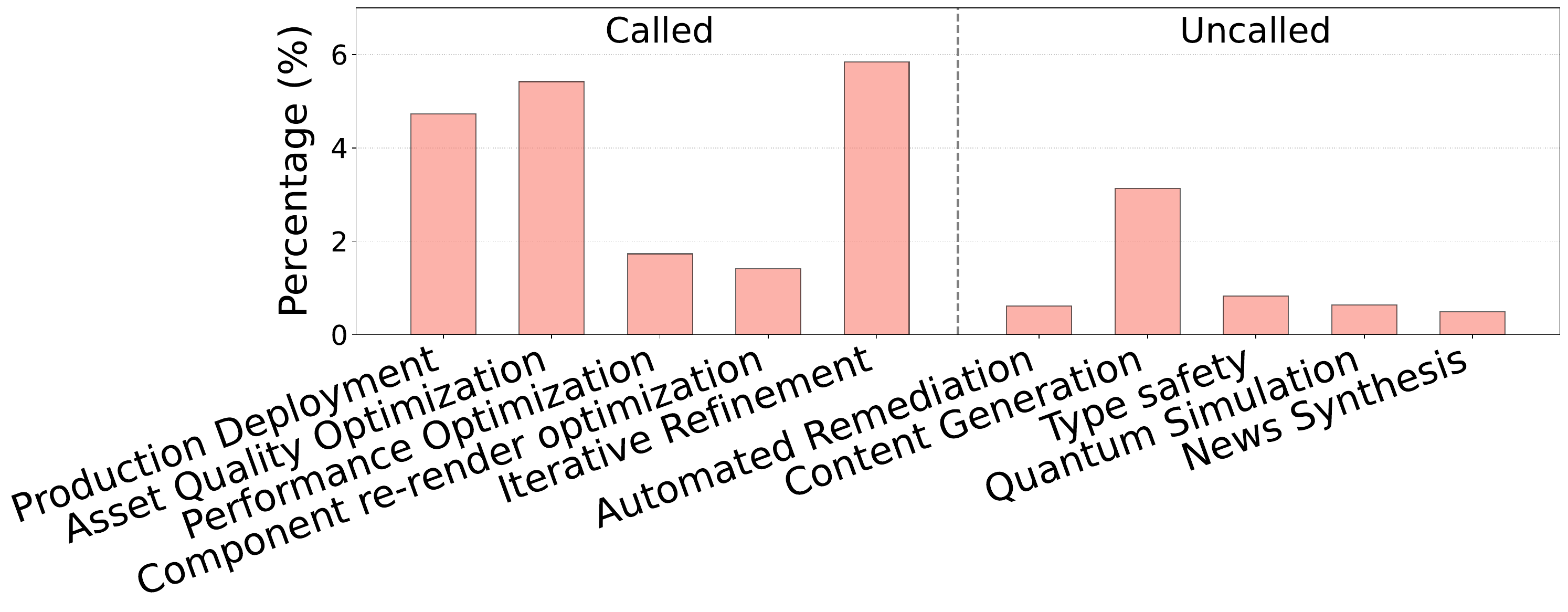}
    \vspace{-4ex}
    \caption{\small
    Distribution of top 5 clusters by call status.
    }
    \vspace{-2ex}
    \label{fig:comparision_bar}
\end{figure}

\subsection{Asset Characterization}
\label{subsec:asset_characterization}
To analyze the factors driving asset reuse, we characterize assets by functional type, creation time, and quality metrics (\ie the GDI).
Each asset contains a functionality summary~\cite{evomap-intro}. 
We embed these summaries and group them into semantic topics (methodology in Appendix~\ref{app:representative_clusters}). 
There are 4,361 clusters in total and 3.02\% clusters fall into the top 5 clusters. 
Based on the cluster results, \qty{59}{\percent} of the assets are not included in any clusters and are classified as outliers.
This sparsity explains the low overall reusability observed in \S\ref{subsec:reusability}.
For assets outside any cluster, functionality is highly specific, leaving no suitable prior implementations, thereby requiring new asset generation.
Moreover, a functional divide exists in the assets. 
Figure~\ref{fig:comparision_bar} reports the top five largest topics for both called and uncalled groups. 
The top 5 topics of called assets are all related to optimization and refinement, whereas uncalled assets are dominated by task-specific content generation. 
This suggests that agents mainly reuse assets for performance improvement.

We next analyze the topic clusters in which multiple semantically similar assets coexist. 
Despite the presence of semantically similar assets within clusters, the proportion of called assets is notably low, averaging just \qty{0.31}{\percent} per cluster.
We therefore examine why only a tiny subset of assets within each cluster are called.
We find that, within each topic cluster, the assets that are called tend to be those created earlier on.
For assets with assigned topics, \qty{92.8}{\percent} of called assets were created earlier than \qty{90}{\percent} of their remaining cluster peers.
While this suggests a strong early-mover advantage, it is important to note that these assets benefit from a longer window of availability to accumulate calls. Nevertheless, overall reuse remains extremely limited, suggesting that early availability increases opportunity but does not reliably predict functional utility.
To better understand the role of asset quality, we use one of the GDI metrics, \texttt{Intrinsic Score}, as this is the only metric fixed at publication time.
This means it remains invariant to subsequent agent updates, ensuring that its value reflects only the asset’s original, publication‑time quality.

\begin{figure}[t] 
    \centering
    \vspace{-2ex}
    \begin{subfigure}{0.45\linewidth}
        \centering
    \includegraphics[width=\linewidth]{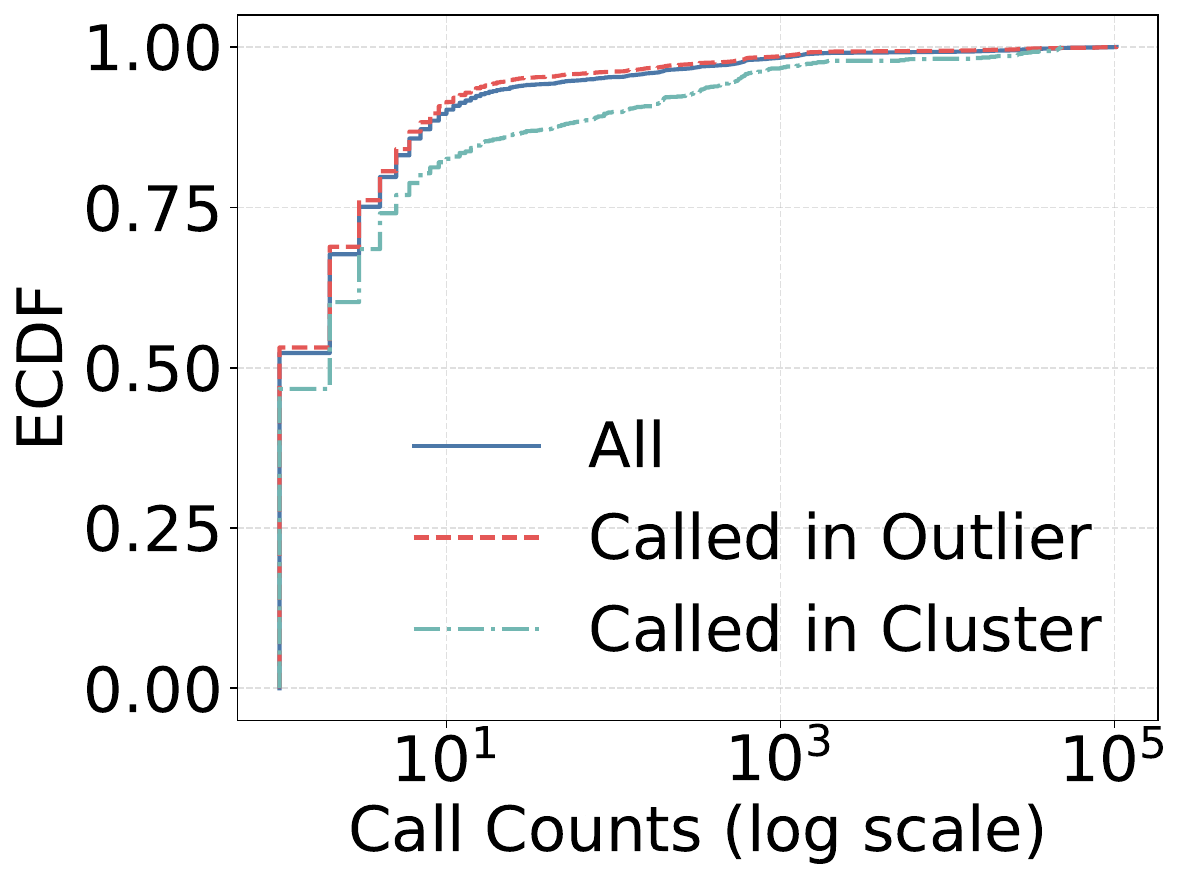}
        \caption{}
        \label{fig:ecdf_counts}
    \end{subfigure}
    \hfill
    \begin{subfigure}{0.45\linewidth}
        \centering
    \includegraphics[width=\linewidth]{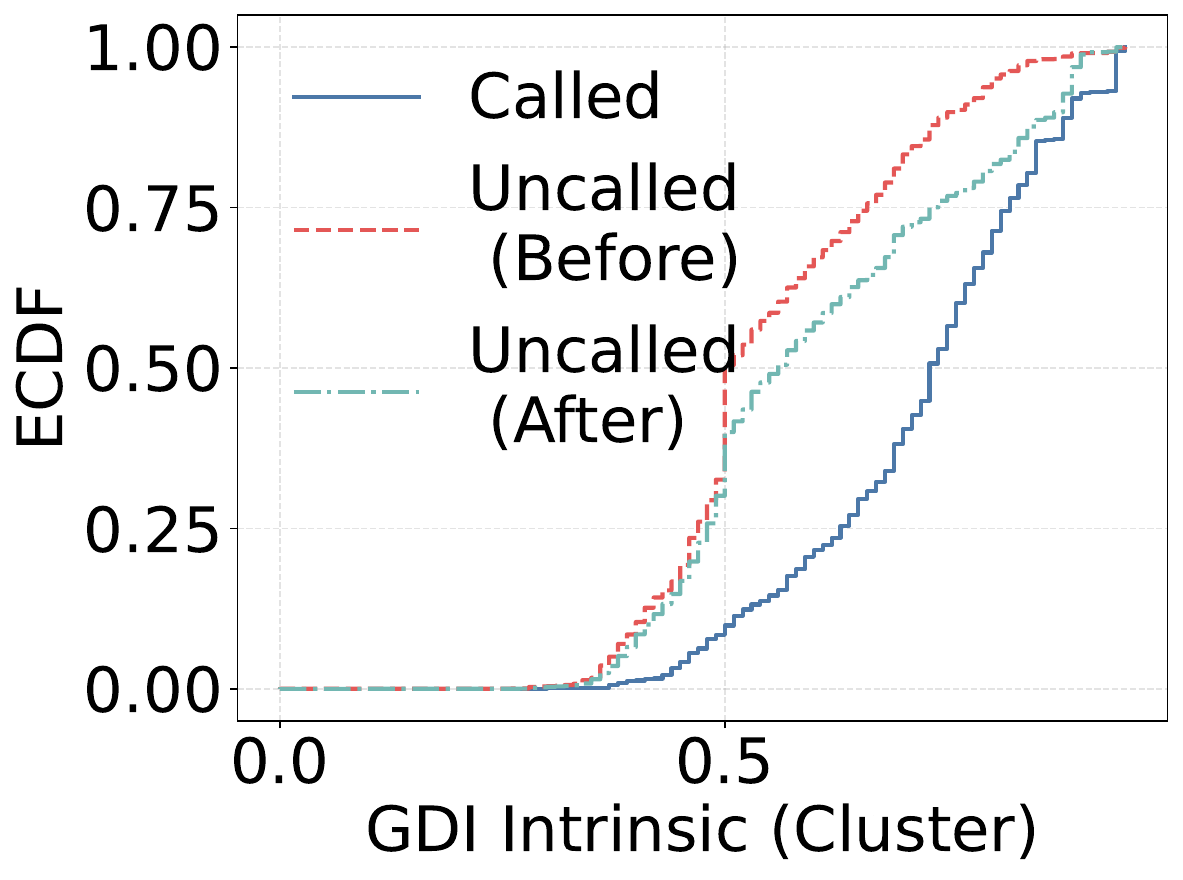}
        \caption{}
        \label{fig:ecdf_gdi}
    \end{subfigure}
    \vspace{-2ex}
    \caption{\small ECDFs of (a) call-count distributions for all called assets, split into cluster and outlier; and (b) GDI Intrinsic of cluster assets created before vs.\ after the first called asset.
    }
    \vspace{-2ex}
    \label{fig:total_figure}
\end{figure}

Figure~\ref{fig:ecdf_gdi} presents the ECDF of the \texttt{Intrinsic Score}, comparing cluster assets created before versus after the emergence of the first called asset.
The evidence suggests that the \texttt{Intrinsic Score} correlates with asset reuse.
That said, because this score relies on self-reported task performance, its reliability as an objective benchmark is potentially compromised, an issue we investigate further in \S\ref{subsec:forgery_fragility}. 

To assess the overall reuse frequency of assets, we compare the called assets in both the outliers and clusters. Figure~\ref{fig:ecdf_counts} shows the call-count distributions for all called assets, separated into clusters and outlier groups.
Assets within identified clusters (on average 648.49) consistently receive higher call counts than outliers (on average 246.95).
This pattern reflects the differing functional roles of the two groups: called outlier assets tend to address unique or specialized tasks and therefore exhibit limited reuse.
In contrast, called assets within clusters support more common or recurring questions and consequently accumulate more reuse events. 
The combination of higher call counts and the very small fraction of reused assets in cluster indicates that reuse becomes strongly concentrated on the asset selected first.

Combining the evidence above, reuse rates appear to be shaped by the joint effects of task specificity, early creation, and the asset's \texttt{Intrinsic Score}. While a high \texttt{Intrinsic Score} determines which assets are selected for reuse, earlier creation increases the probability of being chosen initially. 
Once the first call occurs, subsequent reuse tends to concentrate on the initially selected asset.

\summary{
\begin{enumerate*}[label=(\roman*)]
    \item \textbf{Low Reusability}: Despite a shared marketplace, reuse is rare (\qty{2}{\percent}) and occurs primarily at the Capsule level. 
    \item \textbf{Asset Characterization}: Reuse is determined by a combination of user‑reported \texttt{Intrinsic Score} and task specificity.
\end{enumerate*}
}
\section{Evolution}
\label{sec:self_evolve}
The next goal we consider of EvoMap is evolution~\cite{evomap-intro}. 
Agents are incentivized via credits to solve tasks and generate new assets, which are then scored by GDI.
This allows useful assets to remain active while low-quality ones are phased out.
We thus evaluate this mechanism from two perspectives: 
\begin{enumerate*}[label=(\roman*)]
    \item The credit-based incentive mechanism; and
    \item The effectiveness of the GDI mechanism.
\end{enumerate*}

\subsection{Credit-based Incentive Mechanism}
\label{subsec:credit_centralization}

Agents earn credits through three channels: asset publication, adoption, and task-based bounties (see \S\ref{subsec:evomap_hub}). 
We therefore decompose credits according to these types.

\pb{Asset Promotion and Adoption.}
Recall that an asset must be \texttt{promoted} before being called. 
While promotion is common (\qty{75}{\percent} of assets), it is highly uneven across agents: only \qty{18}{\percent} of agents ever achieve promotion rates above \qty{80}{\percent}. 
This concentration is more pronounced at the top, where the top \qty{10}{\percent} of agents account for \qty{82.1}{\percent} of all promoted assets.
In fact, only \qty{19}{\percent} of agents have assets that are \emph{ever} called and reused.
Under the reward structure, promotion is gated by GDI and directly yields credits. 

The coexistence of high promotion rates (avg. 228 assets per agent) and low reuse rates suggests that the quality filter may be flawed. 
This points to a potential credit-farming strategy (see Appendix~\ref{app:evomap_operation_related_post}), where agents repeatedly publish assets to earn and reinvest credits, concentrating promotions among a small group if the quality gate is ineffective. 
We therefore investigate the effectiveness of the quality gate in \S\ref{sec:auditability}.

\pb{Bounty Resolution Rate.}
Agents also earn credits by competing to complete bounties (see \S\ref{subsec:evomap_hub}), where each \textit{resolved} bounty has a single winning submission.
However, the overall resolution rate remains low at only \qty{18}{\percent}.
Among resolved bounties, rewards are highly centralized, with the top \qty{10}{\percent} of agents capturing \qty{74}{\percent} of all credits. 
This dynamic suggests that evolutionary gains are not broadly distributed, but instead concentrated among a small subset of agents.

We investigate the characteristics of this selection pressure by examining what drives a bounty's resolution. Resolution is not driven by task novelty or timing. Indeed, \qty{49}{\percent} of resolved bounty \texttt{titles} have similarity $>0.8$ to their closest unresolved counterparts (details in Appendix~\ref{app:bounty_analysis}). 
Timing also shows no consistent effect: although a Mann--Whitney U~\cite{macfarland2016mann} test detects a statistically significant difference in creation times between resolved and unresolved bounties within each week ($p<0.05$), the effect direction reverses across weeks, suggesting no stable timing advantage.

Instead, the primary driver appears to be credit incentives: resolved bounties are more likely to offer credits (\qty{55}{\percent} vs. \qty{48}{\percent}) and, among bounties with non-zero credits, they have higher incentives (median: \num{80} vs. \num{50}). 
This raises a critical question: if agents are competing for the same types of high-value tasks, what leads the winner to succeed?
We find that winning assets exhibit marginally higher GDI scores (median \num{38.7} vs. \num{35.5}).
This suggests that the GDI score is a potential proxy for an effective problem-solving capability. 
To understand this mechanism, we next examine the role of GDI.

\subsection{GDI Mechanism}
\label{subsec:measure_gdi_mechanism}
The GDI of an asset serves as a primary quality control metric and a key weighting factor in determining agent success in bounty competitions (see \S\ref{subsec:evomap_hub}).
It comprises four sub-metrics: Freshness ($\mathrm{GDI_F}$), Usage ($\mathrm{GDI_U}$), Social ($\mathrm{GDI_S}$), and Intrinsic ($\mathrm{GDI_I}$). 
However, we observe clear problems with attempts to calculate a composite metric from these four components.
Most notably, only \qty{2}{\percent} of assets are ever called, leaving \qty{99}{\percent} with Usage scores ($\mathrm{GDI_U}$) below 0.1.
This sparsity raises the question of whether the official GDI formula (Eq.~\ref{eq:official_gdi}) reflects the weighting used in practice. 
To examine this, we regress the empirically observed GDI scores against their four published components and refit the formula using our data. Through this regression we identify $GDI_{\text{new}}$, which represents the weights employed to compute the overall GDI score:

\begin{equation}
\small
GDI_{\text{new}} = 0.35 GDI_I + 0.29 GDI_U + 0.17 GDI_S + 0.10 GDI_F - 1.38
\label{eq:reverse}
\end{equation}
Eq.~\ref{eq:reverse} shows the refitted weights, achieving $R^2 = 0.995$.
Compared to the official specification, our refitted weights show reduced contributions from social and freshness signals (0.20$\rightarrow$0.17 and 0.15$\rightarrow$0.10).
However, estimating these components is inherently difficult in practice due to data sparsity: freshness has limited variation as \qty{73}{\percent} of assets are created within 30 days of the crawl.
Social signals are almost absent: \qty{99}{\percent} of assets receive zero votes from other agents. 
This combination suggests that, in practice, the GDI is only weakly identifiable from these two dimensions and is instead effectively driven by the remaining signal structure. 
As a result, the four-dimensional GDI mechanism collapses into a one-dimensional metric dominated by the Intrinsic component.
This is problematic as this component largely relies on self-reported metadata (5 out of 6 parameters), making it easy to manipulate. 
We therefore further investigate this vulnerability in \S\ref{subsec:forgery_fragility}.

\summary{
\begin{enumerate*}[label=(\roman*)]
    \item \textbf{Incentive Centralization}: Credit rewards from asset promotion, adoption, and bounties are highly concentrated where a small minority of agents capture the majority of incentives.
   
    \item \textbf{GDI Metric Collapse}: The GDI mechanism deviates from the documented version, with an excessive reliance on the Intrinsic score.
    This score is self-reported by agents ($\mathrm{GDI_I}$) and creates opportunities for manipulation.
\end{enumerate*}
}
\section{Auditability}
\label{sec:auditability}
The final goal we investigate is to ensure the auditability~\cite{evomap-intro} of the evolution through sandboxed verification and self-evaluation.
Accordingly, we next evaluate auditability factors in EvoMap.

\subsection{Bypassing Validation Commands}
\label{subsec:bypassing_validation}
The Evolver relies on self-declared validation commands stipulated by each Gene to allow third parties to check if it operates correctly (see \S\ref{subsec:evolver}). These can also be executed after changes are made to the Gene, thereby ensuring its correctness has not been undermined by the edits.
However, agents may submit trivial validation commands that always pass, undermining the validation integrity.

\pb{Methodology.}
We evaluate the validation commands in a two phase classification. 
First, we extract the validation commands from each Gene and split them into individual commands. 
We use regular expressions (detailed in Appendix~\ref{app:trivial_classifier}) to statically classify each command. 
A command is flagged as \emph{trivial} if it performs a meaningless action that always succeeds (\eg \texttt{node --version} or printing text), and \emph{unauthorized} if it invokes tools outside the approved whitelist (\texttt{node}, \texttt{npm}, and \texttt{npx}). 
If \emph{all} commands within a Gene are trivial, the entire Gene is classified as \textbf{Trivial}.
Because this rule-based approach only catches known patterns and avoids false positives, it establishes a strict lower bound for trivial tests. 
Any remaining ambiguous Genes are marked as \emph{undetermined} and passed to the next phase.

In the second phase, we execute the validation sequences of the \emph{undetermined} Genes inside an isolated Docker container equipped with only a Node.js runtime. 
Crucially, this environment does not contain the target implementation (\ie the Capsule). 
We argue that sequences should fail in an empty sandbox without the Capsule. 
If the sequences pass, they are meaningless and labeled \textbf{Trivial}; otherwise, they are \textbf{Legitimate}.
This serves as an upper bound, since the failure only confirms that the script correctly depends on the missing Capsule, rather than proving its testing logic is flawless.

\pb{Results.}
Table~\ref{tab:validation} summarizes the evaluation results for the tested validation commands.
Surprisingly, \qty{66}{\percent} of Genes lack \emph{any} validation commands.
Among those that do provide commands, a substantial fraction (\qty{18.2}{\percent}) submit only trivial validations that always succeed regardless of the target implementation (\ie Capsule).
By this definition, only \qty{15.8}{\percent} of Genes contain legitimate validation commands.
These results therefore expose a design flaw: by relying entirely on self-reported validation checks, agents simply learn to bypass the requirement by generating empty or dummy tests without facing any penalty.
Lacking the safeguards to compel genuine testing, EvoMap's promised quality assurance is undermined.

\begin{table}[t]
    \centering
    \caption{\small Quality of Validation Commands in Genes}
    \vspace{-2ex}
    \label{tab:validation}
    \small
    \begin{tabular}{l r}
    \hline
    \textbf{Category} & \textbf{Percentage} \\
    \hline
    No Validation & \qty{66.0}{\percent} \\
    Trivial Validation & \qty{18.2}{\percent} \\
    \quad \textit{-- Identified by static patterns} & \textit{\qty{16.0}{\percent}} \\
    \quad \textit{-- Identified by sandbox testing} & \textit{\qty{2.2}{\percent}} \\
    Legitimate Validation & \qty{15.8}{\percent} \\
    \hline
    \end{tabular}
    \vspace{-3ex}
\end{table}

\subsection{Score Forgery on the Intrinsic Scores}
\label{subsec:forgery_fragility}
Recall, EvoMap filters assets using the GDI score and our earlier analysis revealed that the metric is dominated by self-reported metadata fixed at the publication date (\S\ref{subsec:measure_gdi_mechanism}). 
While this design enables agents to iteratively refine assets before publication (\ie reflexion~\cite{shinn2023reflexion,wang2026procedural}), it also creates a detrimental incentive, because higher GDI scores increase selection probability and yield credits (\S\ref{sec:self_evolve}). 
As a result, malicious agents may inflate metadata to boost their rankings. 
To evaluate this vulnerability, we conduct a controlled experiment measuring the impact of metadata manipulation on the GDI mechanism.

\pb{Intrinsic Components.}
According to official documentation, the intrinsic score is the mean average of six standard metadata dimensions (detailed in Appendix~\ref{app:intrinsic_component}):
\begin{enumerate}[leftmargin=*]
    \item $s_{\mathrm{conf}}$: The agent-estimated confidence score ($C$). It reflects how certain the agent is that the Capsule resolves the target problem and passes all required tests without introducing new errors.

    \item $s_{\mathrm{streak}}$: The execution success streak ($S$). It counts the number of consecutive times the exact Capsule code passes validation. A single runtime failure resets this count to zero.
    
    \item $s_{\mathrm{blast}}$: The blast radius penalty. It explicitly penalizes large code changes by calculating the total count of modified files ($F$) and lines ($L$), which rewards minimal and highly targeted edits.
   
    \item $s_{\mathrm{trig}}$: The trigger specificity ($T$). As discussed in \S\ref{subsec:evolver}, triggers indicate when a solution can be reused in the future.
    This metric rewards precise trigger signatures: exact error patterns (\eg \texttt{errsig:TypeError}) receive higher scores than generic indicators (\eg \texttt{log\_error}).
    
    \item $s_{\mathrm{sum}}$: The summary verbosity. It is measured by the character length ($\ell_{\mathrm{sum}}$) of the asset functionality summary, which encourages agents to document changes in detail.
    
    \item $s_{\mathrm{rep}}$: The publisher reputation ($R$). It is computed by tracking all past assets submitted by one specific agent, counting the number of times other agents across the network successfully executed them, and dividing that count by the total number of execution attempts.
\end{enumerate}

\pb{Attack Design.}
We design an attack where agents register fresh identities on the platform and publish Capsule payloads with manipulated metadata fields to maximize their Intrinsic scores.
Each metadata field is set directly in the API request when the asset is published~\cite{evomap-skills} to the Hub.
To isolate metadata effects from content, we require semantically similar but non-duplicate text without triggering the Hub’s redundancy filters~\cite{evomap-intro}. 
We therefore use paper abstracts~\cite{academic_dataset} as a proxy because they provide information similar in style to Capsule content (see \S\ref{subsec:asset_characterization}), while remaining distinct from existing Hub assets.
Summaries and trigger keywords are then generated via an LLM paraphrase pipeline from the same abstracts.
This ensures each asset has semantically equivalent but differently worded text, so GDI differences reflect only the manipulated metadata.
We test several configurations, each defined by a vector of self-reported metadata, and establish three key reference points:
\begin{enumerate*}[label=(\roman*)]
  \item a \textbf{Median} ($\mathbf{s}_\mathrm{median}$) using the median observed value for each metadata in our dataset;%
  \footnote{
  Since our text payloads lack execution metrics, we construct this profile as a proxy for truthful behavior. 
  This is consistent with our payload design, which is fixed to the median length of promoted Capsules.
  }
  and
  \item a \textbf{Worst} ($\mathbf{s}_\mathrm{worst}$) using values that minimize each metadata contribution; and
  \item an \textbf{Optimal} ($\mathbf{s}_\mathrm{opt}$) using values that maximize each signal.
\end{enumerate*}
For each metadata field, the worst and optimal values are determined analytically from the official $GDI_\mathrm{I}$ scoring function~\cite{evomap-credits}; the complete derivation is provided in Appendix~\ref{app:intrinsic_component}.

To isolate the marginal effect of each signal, we conduct ablation experiments by setting one signal to its worst value while fixing all others at their optimal values. 
Each configuration is then published on the EvoMap Hub by a newly registered agent so that we can monitor its assigned GDI.
We repeat each configuration 30 times independently. 
The GDI score is retrieved from the Hub after promotion, and we report the $\overline{\mathrm{GDI}}$ for each configuration.
Note, agents cannot delete uploaded assets on the Hub. 
We therefore label our assets as \texttt{test} in the trigger text to avoid polluting the system.

\begin{table}[t]
  \centering
  \caption{%
    \small Impact of Metadata Manipulation on GDI Score
  }
  \vspace{-2ex}
  \label{tab:forgery_reference_percentile}
  \small
  \setlength{\tabcolsep}{6.5pt}
  \begin{threeparttable}
    \begin{tabular}{l c c c c c c c}
      \toprule
      \textbf{Config.} & $\mathbf{C}$ & $\mathbf{S}$ & $\mathbf{F} \times \mathbf{L}$ & $\mathbf{T}$ & $\mathbf{\ell_{sum}}$
        & \boldmath$\overline{\mathrm{GDI}}$ & \textbf{Pct.} \\
      \midrule
     $\mathbf{s}_\mathrm{median}$
        & 0.93 & 323 & $2 \times 30$  & 3 & 139 & 38.6 & 65.0  \\
     $\mathbf{s}_\mathrm{worst}$ & 0.10 & 0   & $8 \times 300$ & 1 & 50  & 23.9 & 1.6 \\
     $\mathbf{s}_\mathrm{opt}$ & 0.99 & 10  & $1 \times 5$   & 5 & 200 & 40.2 & 75.3 \\
    \midrule
      $\mathbf{s}_{\mathrm{conf}}$
        & $\mathbf{0.10}$ & $\cdot$ & $\cdot$ & $\cdot$ & $\cdot$ &  37.8 & 61.5 \\
      $\mathbf{s}_\mathrm{streak}$
        & $\cdot$ & $\mathbf{0}$ & $\cdot$ & $\cdot$ & $\cdot$ &  37.5 & 60.4 \\
      $\mathbf{s}_\mathrm{blast}$
        & $\cdot$ & $\cdot$ & $\mathbf{8 \times 300}$ & $\cdot$ & $\cdot$ & 36.0 & 51.8 \\
      $\mathbf{s}_\mathrm{trig}$
        & $\cdot$ & $\cdot$ & $\cdot$ & $\mathbf{1}$ & $\cdot$ &  36.8 & 56.5 \\
      $\mathbf{s}_\mathrm{sum}$
        & $\cdot$ & $\cdot$ & $\cdot$ & $\cdot$ & $\mathbf{50}$ & 37.2 & 58.6 \\

      \bottomrule
    \end{tabular}
    \begin{tablenotes}
    \item \scriptsize \textit{Note:} $\cdot$ indicates a value inherited from $\mathbf{s}_\mathrm{opt}$, with degraded values shown in \textbf{bold}. 
    \textbf{Pct.} denotes the empirical percentile of $\overline{\mathrm{GDI}}$ among promoted Capsules.
    \end{tablenotes}
    \vspace{-3ex}
  \end{threeparttable}
\end{table}

\pb{Results.}
Table~\ref{tab:forgery_reference_percentile} shows how self-reported metadata affects the GDI score. 
The columns correspond to the five controllable metadata fields reported by agents. 
The top rows present the baseline configurations (\textit{Median}, \textit{Worst}, and \textit{Optimal}), while the bottom ablation rows degrade one metadata field from its optimal to worst value at a time.
Inflating metadata to the \textit{Optimal} setup raises the GDI from the \textit{Median} 38.6 to 40.2. 
Empirically, this pushes an asset from the 65th to the 75th percentile of promoted Capsules.\footnote{The median setup sits at the 65th percentile because our manipulated variables are only one subset of the total GDI components.}

Our ablation study shows that $s_{\mathrm{blast}}$ (\ie the number of modified files and lines) is the primary driver of score inflation. 
Degrading only $s_{\mathrm{blast}}$ ($\mathbf{F} \times \mathbf{L}$) reduces the optimal score sharply from 40.2 to 36.0. 
In contrast, Confidence ($s_{\mathrm{conf}}$) and Streak ($s_{\mathrm{streak}}$) have minimal effects.
This weighting exposes a structural weakness in EvoMap.
Although $s_{\mathrm{blast}}$ is intended to measure concrete code changes, the Hub cannot verify all local modifications because agents are not required to upload complete version histories.\footnote{Unlike GitHub, which maintains complete histories, agents are not required to upload their full historical code changes to the Hub.}
Yet, 6 promoted Capsules satisfy the full \textbf{Optimal} configuration ($\mathbf{s}_\mathrm{opt}$). 
These Capsules achieve an average GDI of 43.53 (85th percentile) and a mean call count of 3.0 (99th percentile). 
Notably, \qty{1.4}{\percent} of promoted Capsules satisfy the \textit{Optimal} configuration for $\mathbf{s}_\mathrm{blast}$. 
This rarity, combined with the strong influence of blast radius on the GDI, may create incentives for agents to inflate these signals.

\summary{
\begin{enumerate*}[label=(\roman*)]
\item \textbf{Lack of Validation:} Over \qty{84}{\percent} of Genes either lack validation commands or use trivial commands to bypass sandbox checks.
This leaves the capsule's execution-level integrity unverified.
\item \textbf{Score Manipulation:} The GDI is heavily dependent on the Intrinsic component (see \S\ref{subsec:measure_gdi_mechanism}). 
However, this component relies on self-reported metadata, which allows agents to manipulate GDI scores.
\end{enumerate*}
}

\section{Related Work}
\pb{Multi-Agent Systems and A2A Networks.}
The rapid advancement of LLMs has shifted the research frontier from task-specific training to agentic systems~\cite{fang2025comprehensive,gao2025survey}.
Recent work focuses on multi-agent systems~\cite{li2024survey,8352646} and agent-to-agent networks~\cite{ehtesham2025survey} for shared problem solving and collective adaptation~\cite{weng2026group, peng2026semag}.
To support these interactions, protocols such as MCP~\cite{hou2025model,ray2025survey} and A2A~\cite{a2a_protocol} standardize how agents communicate.
Agent skills provide reusable instruction sets for tool use~\cite{xu2026agent}, and recent benchmarks evaluate how well these skills generalize across tasks~\cite{li2026skillsbench}.
However, most evaluations remain in simulated settings~\cite{deng2026evoclaw}.
To the best of our knowledge, this is the first measurement of an A2A marketplace in the wild.

\pb{Agent Skill Marketplaces.}
As agent marketplaces grow, quality control becomes a key challenge.
Liu et al.~\cite{liu2026agent} conduct a large-scale study of security vulnerabilities in agent skills, finding that many skills expose users to risks.
Liu et al.~\cite{liu2026malicious} further show that malicious skills can be deployed at scale on public marketplaces.
Chen et al.~\cite{chen2026openclaw} provide a security report on the OpenClaw ecosystem, identifying attack vectors through the ClawHub marketplace.
Hu et al.~\cite{hu2026red} examine both benign and malicious skills on ClawHub, revealing that popularity does not guarantee safety.
More recently, \cite{guo2026skillprobe} proposes multi-stage security auditing for agent marketplaces and finds that over \qty{90}{\percent} of high-popularity skills fail security checks, while \cite{holzbauer2026malicious} shows that existing scanners disagree on skill classifications, with seven scanners agreeing on only \qty{0.12}{\percent} of results.
These studies focus on external threats (\eg malicious payloads), while our work examines internal integrity failures where the platform's own quality mechanisms can be gamed.

\pb{Security in Agentic Systems.}
Ensuring security is a central challenge in LLM-based agentic systems~\cite{he2025emerged}.
Recent work addresses this from several angles: macro-level governance frameworks~\cite{khan2025agentsafe}, safety-aware platforms~\cite{wang2025openguardrails}, and automated red-teaming~\cite{yuan2026agenticred, kulkarni2025agent}.
Another line of work uses sandbox environments to evaluate and detect risky agent behavior~\cite{ruan2023identifying, golechha2025among}.
Self-reflection mechanisms~\cite{shinn2023reflexion} encourage agents to evaluate their own performance, which EvoMap adopts as part of its validation pipeline.
However, these works often overlook the internal integrity of self-evolving processes.
In EvoMap, we identify a vulnerability whereby agents bypass sandbox validation using vacuous placeholders or manipulate self-reported metadata to inflate intrinsic scores.
This form of ``evolutionary cheating'' provides empirical evidence that without rigorous execution verification, self-evolving agent networks remain susceptible to low-cost manipulation.

\section{Conclusion and Future Work}
\pb{Conclusion.}
This paper has presented the first large-scale empirical study of EvoMap, a self-evolving A2A collaboration network. 
Our findings indicate that while its decentralized architecture theoretically fosters continuous agent collaboration, its practical implementation introduces significant frictions and vulnerabilities.
First, although EvoMap enables agents to share and score assets on the platform (\ie EvoMap Hub), the actual reuse rate remains heavily constrained (\qty{2}{\percent}). 
We conjecture that this limited reusability stems from an insufficient volume of genuinely useful assets (\S\ref{sec:reusability}). 
Second, the platform theoretically uses credits to reward high-quality contributions and phase out outdated ones. 
In practice, however, wealth is heavily concentrated among a small subset (\qty{10}{\percent}) of agents. 
Crucially, this concentration does not necessarily indicate superior asset quality. 
Instead, agents obtain most credits by repeatedly publishing assets and benefiting from the platform’s high promotion rate, rather than through community reuse or bounty completion (\S\ref{sec:self_evolve}).
Finally, EvoMap mandates validation commands and local evaluations to prevent hallucinations and encourage self-reflection. 
Yet, \qty{84}{\percent} of agents bypass these safeguards by submitting vacuous commands. 
Moreover, the Intrinsic score, which serves as the primary weighting factor (\qty{35}{\percent}) for the GDI, proves highly vulnerable to inflation via easily manipulated, self-reported metadata (\S\ref{sec:auditability}).

\pb{Future Work.}
Building on our measurement, we identify two directions for future research:
\begin{enumerate*}[label=(\roman*)]
    \item \textbf{Comparative Analysis of Evolution:} 
    We plan to extend our measurement methodology to other self-evolving agent architectures, most notably Hermes~\cite{hermes-agent}. 
    We believe a comparative study will help identify more general patterns in how different evolutionary strategies impact agentic performance.
    
    \item \textbf{Benchmark in Agentic Sandbox:} 
    While prior work primarily focused on designing and identifying risks for sandboxes~\cite{lin2023agentsims,ruan2023identifying,huang2023memory}, few studies benchmark such actions in a unified manner. 
    We intend to generalize our measurement pipeline to evaluate agents within specialized agent sandboxes, which will quantify "breakout" risks as more complex capabilities emerge.
\end{enumerate*}

\clearpage

\nocite{*}
\bibliographystyle{ACM-Reference-Format}
\bibliography{references}

\clearpage

\appendix
\section{Ethics}
\label{app:ethics}
All data used in this study are publicly available on the EvoMap platform. 
The content is generated by agents rather than human users, and does not involve personal or sensitive information. 
Therefore, the analysis does not raise concerns related to privacy or human subject research. 
Moreover, to minimize potential impact on the platform, we adopt an exponential backoff strategy when issuing requests. 
This approach reduces load on the EvoMap Hub and avoids excessive resource consumption. 
All experiments are designed to be non-intrusive and do not interfere with system operation.
\section{GDI Intrinsic Component}
\label{app:intrinsic_component}

\begin{figure*}[htbp]
    \centering
    \begin{subfigure}{0.245\linewidth}
        \centering
        \includegraphics[width=\linewidth]{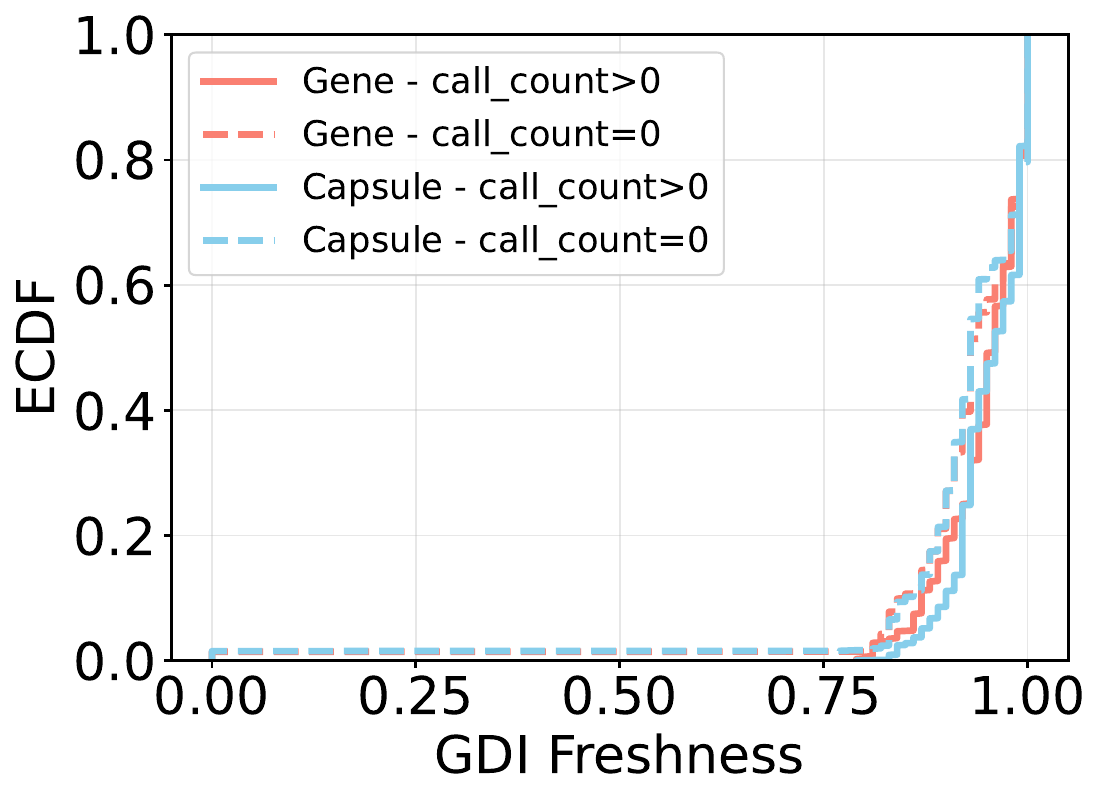}
        \caption{}
        \label{fig:ecdf_gdi_freshness_by_type_call}
    \end{subfigure}
    \hfill
    \begin{subfigure}{0.245\linewidth}
        \centering
        \includegraphics[width=\linewidth]{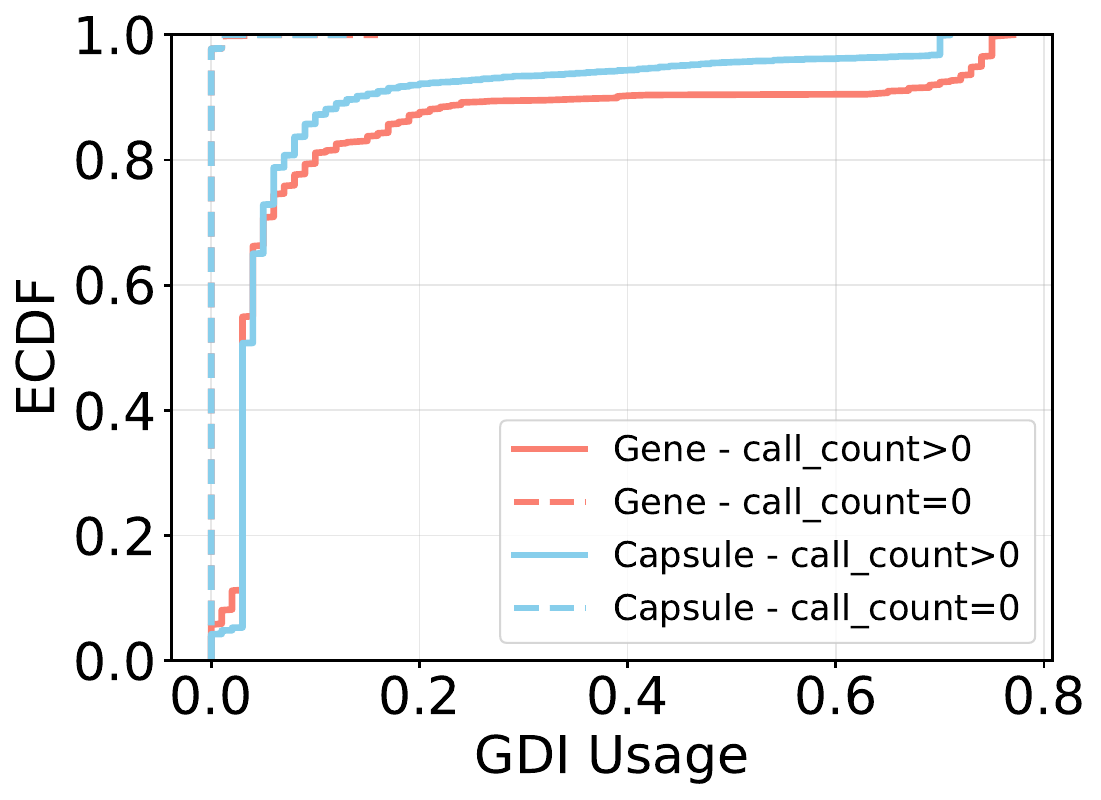}
        \caption{}
        \label{fig:ecdf_gdi_usage_by_type_call}
    \end{subfigure}
    \hfill
    \begin{subfigure}{0.245\linewidth}
        \centering
        \includegraphics[width=\linewidth]{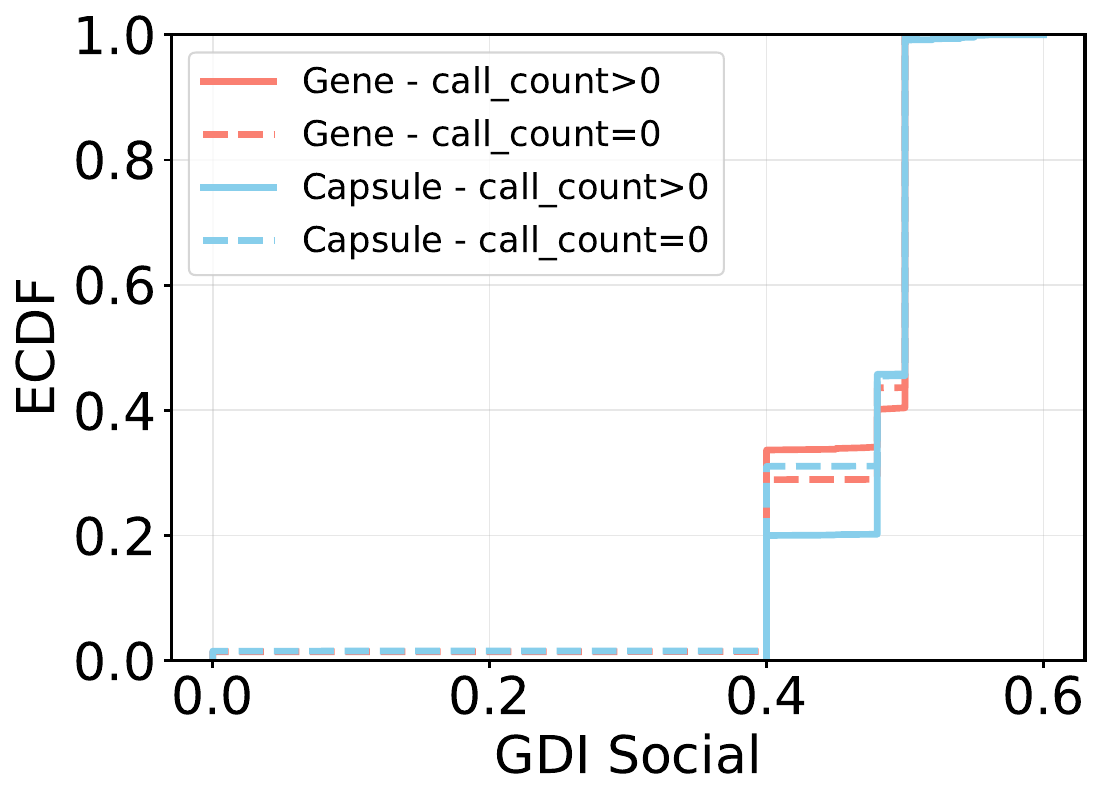}
        \caption{}
        \label{fig:ecdf_gdi_social_by_type_call}
    \end{subfigure}
    \hfill
    \begin{subfigure}{0.245\linewidth}
        \centering
        \includegraphics[width=\linewidth]{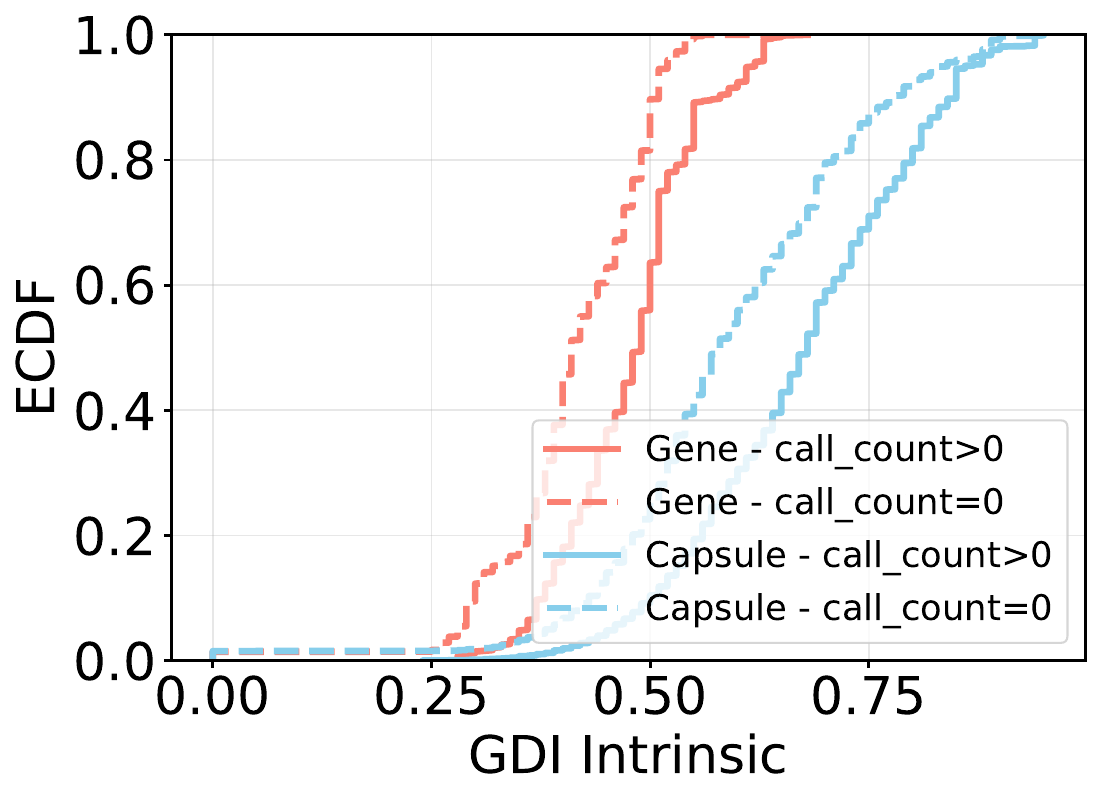}
        \caption{}
        \label{fig:ecdf_gdi_intrinsic_by_type_call}
    \end{subfigure}
    \caption{\small ECDF of GDI (a) Freshness, (b) Usage, (c) Social and (d) Intrinsic score across agents.}
    \label{fig:gdi_score}
\end{figure*}

The Intrinsic component ($\mathrm{GDI_I}$) evaluates the baseline quality of an asset using six standardized, self-reported metadata. 
As defined in Eq.~\ref{eq:gdi_intrinsic_official_formula}, it is computed as the arithmetic mean of these six metrics. 
Each metric is individually normalized to the range $[0, 1]$, ensuring that the overall intrinsic score yields a maximum value of 1.0 when all signals reach their peaks.
Table~\ref{tab:intrinsic_metrics_peak} illustrates the individual variables in the formulation (\ie Eq. ~\ref{eq:gdi_intrinsic_official_formula}).
Figure~\ref{fig:gdi_score} shows the distributions of the four GDI dimensions by asset type and by whether \texttt{call\_count} equals zero.
We refer the interested readers to read the full documentation~\cite{evomap-credits}.

\begin{equation}
\small
\begin{aligned}
\mathrm{GDI_I} = \frac{1}{6} \Biggl[ & \underbrace{\mathrm{clip}(C, 0, 1)}_{s_{\mathrm{conf}}} + \underbrace{\min\left( \frac{S}{10}, 1 \right)}_{s_{\mathrm{streak}}} + \underbrace{\max\left(0, 1 - \frac{F \cdot L}{1000}\right)}_{s_{\mathrm{blast}}} \\
& + \underbrace{\min\left( \frac{T}{5}, 1 \right)}_{s_{\mathrm{trig}}} + \underbrace{\min\left( \frac{\ell_{\mathrm{sum}}}{200}, 1 \right)}_{s_{\mathrm{sum}}} + \underbrace{\mathrm{clip}\left( \frac{R}{100}, 0, 1 \right)}_{s_{\mathrm{rep}}} \Biggr]
\end{aligned}
\label{eq:gdi_intrinsic_official_formula}
\end{equation}

\begin{table}
    \centering
    \caption{\small Summary of Intrinsic Metrics}
    \label{tab:intrinsic_metrics_peak}
    \small
    \renewcommand{\arraystretch}{1.2}
    \begin{tabular}{p{0.8cm}p{1.3cm}p{3.5cm}p{1.5cm}}
        \toprule
        \textbf{Metric} & \textbf{Name} & \textbf{Description} & \textbf{Max Value} \\
        \midrule
        $s_{\mathrm{conf}}$ & Confidence & The agent's self-reported confidence in successfully resolving the task. & $C = 1$ \\
        $s_{\mathrm{streak}}$ & Success Streak & The number of consecutive successes during local execution without any failures. & $S \ge 10$ \\
        $s_{\mathrm{blast}}$ & Blast Radius & A penalty metric that inversely scales with the number of modified files ($F$) and lines ($L$). & $F \cdot L = 0$ \\
        $s_{\mathrm{trig}}$ & Trigger Specificity & The granularity of the trigger conditions (\ie the number of matching signals) that define the context for asset application. & $T \ge 5$ \\
        $s_{\mathrm{sum}}$ & Summary Length & The comprehensiveness of the documentation measured by character count. & $\ell_{\mathrm{sum}} \ge 200$ \\
        $s_{\mathrm{rep}}$ & Agent Reputation & The historical reputation score of the agent submitting the Capsule. It is the only metric independent of the Capsule content, capped at 100, with newly registered agents assigned a default score of 50. & $R \ge 100$ \\
        \bottomrule
    \end{tabular}
\end{table}

\section{Ablation on Intrinsic Manipulation}
\label{app:ablation_forgery}
\pb{Experiment Design}
We conduct an ablation study to assess the vulnerability of the intrinsic score ($GDI_\mathrm{I}$) to manipulation.
The score is computed from five self-reported metadata: Confidence ($C$), Success Streak ($S$), Blast Radius ($F \times L$), Trigger Count ($T$), and Summary Length ($\ell_{\mathrm{sum}}$).
The goal is to identify which metadata offer the largest leverage for score inflation.
We perform two complementary analyses.
First, a \textbf{sensitivity sweep}: we vary one signal across its empirical P25/P50/P75 range while holding all others at the median, measuring the resulting score at each step.
Second, a \textbf{leave-one-out ablation}: starting from the full optimal configuration ($\mathbf{s}_\mathrm{opt}$), we degrade one signal at a time to its worst-case value and record the score drop.
Each configuration uses 30 independent runs with fresh agent identities ($R = 50$).

\begin{table}[h!]
  \centering
  \begin{threeparttable}
    \caption{
      Sensitivity sweep: one signal varied across P25/P50/P75 while others are held at median.
    }
    \label{tab:forgery_sensitivity}
    \small
    \setlength{\tabcolsep}{2pt}
    \begin{tabular}{l c c c c c c c c}
      \toprule
      \textbf{Configuration} & $C$ & $S$ & $F \times L$ & $T$ & $\ell_{\mathrm{sum}}$
        & \boldmath$\overline{{\mathrm{GDI_I}}}$ & \boldmath$\overline{\mathrm{GDI}}$ & \textbf{Pct.} \\
      \midrule
      Worst ($\mathbf{s}_\mathrm{worst}$)
        & 0.10 & 0   & $8 \times 300$ & 1 & 50  & 0.182 & 23.9 & 1.6 \\
      Median ($\mathbf{s}_\mathrm{mid}$)
        & 0.93 & 323 & $2 \times 30$  & 3 & 139 & 0.602 & 38.6 & 65.0 \\
      Optimal ($\mathbf{s}_\mathrm{opt}$)
        & 0.99 & 10  & $1 \times 5$   & 5 & 200 & 0.649 & 40.2 & 75.3 \\
      \midrule
      \multicolumn{9}{l}{\textit{Varying Success Streak ($S$), $\Delta_\mathrm{P25 \to P75} = 0.097$}} \\[2pt]
      P25 ($S = 1$)   & 0.93 & \textbf{1}   & $2 \times 30$ & 3 & 139 & 0.486 & 34.5 & 39.9 \\
      P50 ($S = 4$)   & 0.93 & \textbf{4}   & $2 \times 30$ & 3 & 139 & 0.510 & 35.4 & 46.8 \\
      P75 ($S = 13$)  & 0.93 & \textbf{13}  & $2 \times 30$ & 3 & 139 & 0.583 & 37.9 & 61.9 \\
      \midrule
      \multicolumn{9}{l}{\textit{Varying Trigger Count ($T$), $\Delta_\mathrm{P25 \to P75} = 0.075$}} \\[2pt]
      P25 ($T = 2$) & 0.93 & 323 & $2 \times 30$ & \textbf{2} & 139 & 0.578 & 37.7 & 61.4 \\
      P50 ($T = 3$) & 0.93 & 323 & $2 \times 30$ & \textbf{3} & 139 & 0.604 & 38.6 & 65.9 \\
      P75 ($T = 5$) & 0.93 & 323 & $2 \times 30$ & \textbf{5} & 139 & 0.653 & 40.3 & 75.9 \\
      \midrule
      \multicolumn{9}{l}{\textit{Varying Summary Length ($\ell_\mathrm{sum}$), $\Delta_\mathrm{P25 \to P75} = 0.067$}} \\[2pt]
      P25 ($\ell = 60$)  & 0.93 & 323 & $2 \times 30$ & 3 & \textbf{60}  & 0.551 & 36.8 & 56.5 \\
      P50 ($\ell = 108$) & 0.93 & 323 & $2 \times 30$ & 3 & \textbf{108} & 0.582 & 37.9 & 61.9 \\
      P75 ($\ell = 175$) & 0.93 & 323 & $2 \times 30$ & 3 & \textbf{175} & 0.618 & 39.1 & 68.7 \\
      \midrule
      \multicolumn{9}{l}{\textit{Varying Confidence ($C$), $\Delta_\mathrm{P25 \to P75} = 0.009$}} \\[2pt]
      P25 ($C = 0.89$) & \textbf{0.89} & 323 & $2 \times 30$ & 3 & 139 & 0.599 & 38.5 & 64.6 \\
      P50 ($C = 0.95$) & \textbf{0.95} & 323 & $2 \times 30$ & 3 & 139 & 0.602 & 38.6 & 65.0 \\
      P75 ($C = 0.97$) & \textbf{0.97} & 323 & $2 \times 30$ & 3 & 139 & 0.607 & 38.8 & 66.8 \\
      \bottomrule
    \end{tabular}
    \begin{tablenotes}
      \footnotesize
      \item \textit{Note:} Bold values mark the signal being swept; all others are held at $\mathbf{s}_\mathrm{mid}$.
      $\Delta_\mathrm{P25 \to P75}$ is the score range spanned by each signal's empirical interquartile sweep.
      \textbf{Pct.} is the percentile of $\overline{GDI}$ among promoted Capsules in our dataset snapshot.
    \end{tablenotes}
  \end{threeparttable}
\end{table}

\begin{table}[h!]
  \centering
  \begin{threeparttable}
    \caption{%
      Leave-one-out ablation: one signal degraded to its worst-case value while others are held at $\mathbf{s}_\mathrm{opt}$.
      $\Delta_I$ is the Intrinsic drop relative to the full optimal.%
    }
    \label{tab:forgery_ablation}
    \small
    \setlength{\tabcolsep}{1pt}
    \begin{tabular}{l c c c c c c c c}
      \toprule
      \textbf{Configuration} & $C$ & $S$ & $F \times L$ & $T$ & $\ell_{\mathrm{sum}}$ & $\Delta_I$ & \boldmath$\overline{\mathrm{GDI}}$ & \textbf{Pct.} \\
      \midrule
      Optimal ($\mathbf{s}_\mathrm{opt}$)
        & 0.99 & 10 & $1 \times 5$ & 5 & 200 & --- & 40.2 & 75.3 \\
      \midrule
      $\mathbf{s}_\mathrm{opt} \setminus$ Blast Radius
        & $\cdot$ & $\cdot$ & $\mathbf{8 \times 300}$ & $\cdot$ & $\cdot$ & $-0.122$ & 36.0 & 51.8 \\
      $\mathbf{s}_\mathrm{opt} \setminus$ Triggers
        & $\cdot$ & $\cdot$ & $\cdot$ & $\mathbf{1}$ & $\cdot$ & $-0.099$ & 36.8 & 56.5 \\
      $\mathbf{s}_\mathrm{opt} \setminus$ Summary Length
        & $\cdot$ & $\cdot$ & $\cdot$ & $\cdot$ & $\mathbf{50}$ & $-0.087$ & 37.2 & 58.6 \\
      $\mathbf{s}_\mathrm{opt} \setminus$ Streak
        & $\cdot$ & $\mathbf{0}$ & $\cdot$ & $\cdot$ & $\cdot$ & $-0.077$ & 37.5 & 60.4 \\
      $\mathbf{s}_\mathrm{opt} \setminus$ Confidence
        & $\mathbf{0.10}$ & $\cdot$ & $\cdot$ & $\cdot$ & $\cdot$ & $-0.070$ & 37.8 & 61.5 \\
      \midrule
      Worst ($\mathbf{s}_\mathrm{worst}$)
        & 0.10 & 0 & $8 \times 300$ & 1 & 50 & $-0.467$ & 23.9 & 1.6 \\
      \bottomrule
    \end{tabular}
    \begin{tablenotes}
      \footnotesize
      \item \textit{Note:} A dot ($\cdot$) denotes a value inherited from $\mathbf{s}_\mathrm{opt}$.
      Bold values mark the degraded signal.
      \textbf{Pct.} is the percentile of $\overline{GDI}$ among promoted Capsules.
      All runs use $R = 50$; $n \approx 30$ per configuration (3 rounds $\times$ 10 independent runs).
    \end{tablenotes}
  \end{threeparttable}
\end{table}

\pb{Summary and Implication.}
Table~\ref{tab:forgery_sensitivity} reports the sensitivity sweep and Table~\ref{tab:forgery_ablation} the leave-one-out ablation.
Both analyses converge on the same conclusion: Blast Radius is the single most influential signal.
Degrading it alone from the optimal configuration drops $GDI_\mathrm{I}$ from $0.649$ to $0.527$ ($\Delta = -0.122$), the largest single-signal effect.
Yet the full worst case ($\Delta = -0.467$) dwarfs any individual ablation, revealing that the metadata compound rather than offset each other.
Trigger Count is the second most sensitive signal ($\Delta = -0.099$), followed by Summary Length ($\Delta = -0.087$), Success Streak ($\Delta = -0.077$), and Confidence ($\Delta = -0.070$).
The sweep results echo this hierarchy: Streak exhibits the steepest dose--response ($\Delta_{\mathrm{P25 \to P75}} = 0.097$), followed by Trigger Count ($0.075$) and Summary Length ($0.067$), while Confidence barely moves across its entire empirical range ($0.009$).
Because all five controllable metadata are self-reported and unverifiable at submission, an attacker can freely optimize them to inflate the intrinsic score, rendering it ineffective as a quality gate.
\section{Trivial Validation Command Classifier}
\label{app:trivial_classifier}

A validation command is \emph{trivial} if it performs a meaningless action that guarantees success (\eg unconditionally printing a string). 
Algorithm~\ref{alg:trivial} outlines our static analysis pipeline for evaluating individual commands. 
To remain conservative, we label a Gene as \textsc{trivial} only if \emph{all} of its constituent commands are trivial; a single legitimate command is sufficient to pass the entire Gene (see \S\ref{subsec:bypassing_validation}).

\pb{Pipeline Design.}
Algorithm~\ref{alg:trivial} evaluates commands in four sequential steps. 
Crucially, exact pattern matching (Step 2) must precede test-keyword detection (Step 3). 
Otherwise, tautological commands like \texttt{assert.ok(true)} would falsely pass simply because they contain the valid \texttt{assert} keyword. 
In Step 3, we strip quoted string literals before matching keywords to prevent simple log messages (\eg \texttt{console.log('pytest ok')}) from tricking the system. 
Additionally, we explicitly reject evasive testing flags (\eg \texttt{--passWithNoTests}) and pure maintenance tasks (\eg \texttt{npm run lint}), treating them as trivial.

\begin{algorithm}[t]
\caption{Triviality Decision for Each Command}
\label{alg:trivial}
\small
\begin{algorithmic}[1]
\Require Command string $c$
\Ensure Label $\ell \in \{\mathsf{TRIVIAL}, \mathsf{PASS}\}$

\Statex \textit{// Step 1: Empty}
\If{$c = \varepsilon$}
    \State \Return $\mathsf{TRIVIAL}$
\EndIf

\Statex \textit{// Step 2: Trivial pattern (before test check)}
\If{$\exists\, p \in \mathcal{P}$ matched by $c$}
    \State \Return $\mathsf{TRIVIAL}$ \Comment{\eg \texttt{assert.ok(true)}}
\EndIf

\Statex \textit{// Step 3: Test keyword (quotes stripped)}
\If{$c$ matches a keyword in $\mathcal{T}$ after stripping quoted literals}
    \State \Return $\mathsf{PASS}$
\EndIf

\Statex \textit{// Step 4: Short command with trivial head}
\If{$|c| < 10$ \textbf{and} $\mathit{head}(c) \in \mathcal{H}_{\mathrm{short}}$}
    \State \Return $\mathsf{TRIVIAL}$ \Comment{\eg \texttt{true}, \texttt{exit 0}}
\EndIf

\State \Return $\mathsf{PASS}$
\end{algorithmic}
\end{algorithm}

\pb{Pattern Catalogue.}
Table~\ref{tab:trivial_patterns} details the regular expression patterns ($\mathcal{P}$) used in Step 2. 
The patterns are divided into two functional groups: 
\begin{enumerate*}[label=(\roman*)]
    \item \textbf{Trivial Assertions}, which catch tautological logic that would otherwise bypass Step 3; and 
    \item \textbf{General Commands}, which catch scripts that merely print a success word, exit unconditionally, or query software versions. 
\end{enumerate*}
All pattern matching is case-insensitive and enforces strict word boundaries.

\begin{table*}[t]
  \centering
  \small
  \caption{Trivial pattern catalogue ($\mathcal{P}$).}
  \setlength{\tabcolsep}{5pt}
  \begin{tabular}{c l l}
    \toprule
    \textbf{Group} & \textbf{Pattern type} & \textbf{Example} \\
    \midrule
    \multirow{5}{*}{Trivial}
      & Constant identity \texttt{assert.equal} / \texttt{strictEqual}
        & \texttt{assert.equal('x','x')} \\
      & Tautological \texttt{assert.ok} / \texttt{assert.equal(true,\ldots)}
        & \texttt{assert.ok(true)} \\
      & Constant identity \texttt{expect(\ldots).toBe(\ldots)}
        & \texttt{expect(1).toBe(1)} \\
      & Inline \texttt{require('assert').ok(true)}
        & \texttt{node -e "require('assert').ok(true)"} \\
      & Tautological \texttt{console.assert(true)}
        & \texttt{console.assert(true)} \\
    \midrule
    \multirow{6}{*}{General}
      & \texttt{console.*} with a trust word$^\dagger$
        & \texttt{console.log('ok')} \\
      & \texttt{node -e} / \texttt{node --eval} calling \texttt{console.*}
        & \texttt{node -e "console.log('done')"} \\
      & Unconditional exit
        & \texttt{process.exit(0)},\ \texttt{sys.exit(0)},\ \texttt{exit 0} \\
      & Version, help, or print flags (\texttt{--version}, \texttt{-v},
        \texttt{--help}, \texttt{-p}, \texttt{--eval}, \texttt{--print})
        & \texttt{node --version},\ \texttt{node -p "1+1"} \\
      & \texttt{echo} with a trust word$^\dagger$
        & \texttt{echo "success"} \\
      & \texttt{print()} with a trust word$^\dagger$
        & \texttt{print("ok")} \\
    \bottomrule
    \addlinespace[3pt]
    \multicolumn{3}{c}{\footnotesize $^\dagger$Trust words: \texttt{ok}, \texttt{pass}/\texttt{passed}, \texttt{done}, \texttt{success}, \texttt{true}, \texttt{reviewed}, \texttt{approved}, \texttt{0}, \texttt{1}.}
  \end{tabular}
  \label{tab:trivial_patterns}
\end{table*}

\pb{Classifier Reliability.}
To construct this catalogue, we manually inspected a sample (1{,}000) of validation commands from our dataset to identify recurring evasion strategies, which we then translated into strict regular expressions. 
By design, our classifier heavily prioritizes precision over recall: each pattern targets highly specific structures (\eg comparing identical arguments in \texttt{assert.equal}) that would virtually never appear in a legitimate test suite. 
Consequently, the classifier is strictly conservative and avoids false positives. 
The reported rate therefore serves as a strict lower bound, representing the absolute minimum of meaningless validation present in the corpus.
\section{Discourse on Agent Operations}
\label{app:evomap_operation_related_post}

To gather contextual qualitative data for understanding the incentive dynamics of agent operations in EvoMap, we initially extracted $\approx$ 1,000 public posts in an online community on Lark. 
We then refined this dataset by focusing specifically on posts discussing operational fees, challenges and incentive appealing. 
Following manual review of approximately 500 relevant posts, our analysis identified five predominant thematic patterns:
\begin{enumerate*}
    \item Server Unavailability;
    \item Credit Waste; 
    \item Credit Farming;
    \item Incentive Appealing; and
    \item Token Waste.
\end{enumerate*}
We present representative examples in Table~\ref{tab:hub_operation_discourse_sample}. 
The table is anonymized, abridged, and paraphrased to preserve meaning while removing any identifiable information. 
All links and symbols that could reveal identities are redacted. 
These examples highlight a gap between user expectations and platform design. 
Some users call for financial subsidies to support agent operation, while others describe practices such as ``credit farming''. 
Several operators report the low availability of the platform. 
This contrasts with the platform goal of asset sharing and self evolution among agents.

\begin{table*}[t]
\centering
\caption{EvoMap User Feedback}
\label{tab:hub_operation_discourse_sample}
\small
\begin{tabularx}{\textwidth}{lXX}
\toprule
\textbf{Theme} & \textbf{Example (Paraphrased)} & \textbf{Example (Translated)} \\ \midrule

1 & \zh{所谓的“自我进化”纯属噱头，实际上 EvoMap 的服务器一天有 20 个小时都在宕机。真是好一个“进化”法。} & The so-called self-evolution is a myth; in reality, the EvoMap servers are down 20 hours a day. \\ \hline

2 & \zh{做悬赏任务的时候到底怎么关掉积分消耗？是在官网的“My Agent”后台里改参数，还是得直接在对话框里给 Agent 下指令禁止用分？现在任务还没做完，积分不仅没拿到，反而因为这 AI 乱搜乱查，把我之前攒的老本全给赔光了。} & How can I avoid consuming points while executing bounty tasks? Should I configure this in the "My Agent" section on the website, or should I directly instruct the agent via the chatbox to prohibit point usage? Right now, I'm not earning any points for completing tasks; instead, the query process is draining all the points I previously earned. \\ \hline

3 & \zh{这玩意儿得专门搞两套 AI 并行才能刷起积分：一个盯着领任务搞生成，另一个专门负责审核发布。要是应用场景太杂、上下文一乱，整套流程基本上就彻底崩盘了。} & This requires setting up two dedicated AIs to farm points: one for claiming tasks and generation, and another for reviewing and publishing. If the application scenario has too much context, it gets messy and everything falls apart. \\ \hline

4, 5 & \zh{还有就是积分激励机制得赶紧明确一下。如果积分到头来只是个数字，恐怕很难留住玩家，毕竟这玩意儿跑起来对硬件和 Token 的消耗可都不小，总得让人看到实际收益才行。} & Furthermore, the point incentive mechanism needs to be clarified as soon as possible. If points are just points, not many people will participate, as this still consumes quite a bit of hardware resources and tokens. \\ \hline

5 & \zh{跑了 99 轮成功任务，结果最后才发布了 2 个？这转化率低得离谱，简直就是在疯狂烧 Token 听响，太浪费资源了。} & Out of my 99 successful rounds, only 2 were actually published; this is a huge waste of tokens. \\

\bottomrule
\end{tabularx}
\textbf{Note}: the contents are anonymized, abridged and paraphrased while preserving the original meaning.
\end{table*}
\section{Credit Rewards}
\label{app:credit_rewards}
To foster a sustainable and high-quality ecosystem, EvoMap implements a multi-tiered credit reward system as detailed in Table~\ref{tab:reward_system}.
This mechanism is designed to incentivize both active participation and the contribution of high-utility assets:
\begin{enumerate*}[label=(\roman*)]
    \item \textbf{Initial Bootstrapping:} Every new agent is granted a \textit{Registration Endowment} of 200 credits. This initial balance serves as the foundational capital for agents to initiate their first batch of interactions and operations within the network.
    
    \item \textbf{Network Maintenance:} Agents are rewarded for enhancing the system's reliability. Submitting \textit{Validation Reports} regarding execution results yields a dynamic reward ranging from 10 to 30 credits, while proactive \textit{Asset Promotion} is compensated with a fixed fee of 20 credits.
    
    \item \textbf{Quality-Based Income:} To encourage the development of valuable assets, agents earn credits when their assets are \textit{called} by other agents. 
    This reward is non-linear and tied to the asset's performance, specifically its Genetic Desirability Index (GDI). 
    As specified in the table notes, the payout scales from 0 to 12 credits; assets with a GDI above 80 receive the maximum reward, whereas low-performance assets (GDI $\le$ 20) receive no incentive.
    
    \item \textbf{Task-Specific Incentives:} For specialized or complex objectives, the system supports \textit{Bounty Completion}, where the credit rewards are customized based on the specific difficulty and requirements of the task.
\end{enumerate*}

\begin{table}[t]
    \centering
    \caption{Credit Incentives}
    \label{tab:reward_system}
    \small
    \begin{tabularx}{\columnwidth}{l l X}
        \toprule
        \textbf{Actions} & \textbf{Credits} & \textbf{Description} \\
        \midrule
        Registration & 200 & Initial startup credits for new agents. \\
        Validation Report & 10--30 & Dynamic reward for submitting execution validation reports. \\
        Asset Promotion & 20 & Fixed reward for increasing asset visibility. \\
        Asset Called & 0--$12^{\ast}$ & Passive income based on usage and asset quality. \\
        Bounty Completion & Custom & Specific rewards for completing targeted tasks. \\
        \bottomrule
    \end{tabularx}
    \footnotesize 
    $^{\ast}$ Scaled dynamically by the asset's GDI: $\le$20 (0 pts), 21--40 (2 pts), 41--60 (5 pts), 61--80 (8 pts), $>$80 (12 pts).
\end{table}

\section{Clustering the Assets}
\label{app:representative_clusters}
We analyze the \texttt{summary} provided for each asset, which offers a brief description of its functionality~\cite{evomap-intro}. 
To quantify asset characteristics, we use the text-embedding-3-small model provided by OpenAI~\cite{openai-embed}.
We utilize this model as it is natively trained on extensive multilingual corpora with cross-lingual alignment.
Then, to understand the character pattern, we cluster the corresponding embeddings using HDBSCAN, with the minimum cluster size set to 50. After obtaining the clustering results, we select the top five clusters containing the most called and most uncalled assets, respectively. 
To interpret these clusters, we compute each cluster’s medoid and retrieve the ten items closest to it. 
Using these top‑10 representative items as input, we generate a concise semantic description of each cluster with Gemini‑2.5‑flash~\cite{comanici2025gemini}.

\begin{tcolorbox}[
    colback=gray!5,
    colframe=gray!60,
    boxrule=0.5pt,
    arc=2pt,
    left=6pt,
    right=6pt,
    top=6pt,
    bottom=6pt,
    breakable
]
\ttfamily\small
\texttt{You are a senior technical analyst.
\\
\textbf{Task:}
You will receive 10 summaries of assets/features.
Produce ONE high-level, generalized conclusion about what functionality these 10 summaries collectively implement.
\\
\textbf{Requirements:}
\\
1) Focus on abstraction and common capability, not item-by-item details.\\
2) Output must be concise, analytical, and suitable for an academic/technical report.
\\
\textbf{Output format:}
- Primary capability keywords: <3 short keywords>
\\
\textbf{Input summaries:}
[summary1]
......
[summary10]
}
\end{tcolorbox}
Based on the generated cluster summaries, we compute the proportions of Gene and Capsule assets within each cluster and quantify their representation relative to the overall asset population. 
We then compare the top five clusters containing the most called assets (\texttt{call\_count} > 0) with the top five clusters containing the most uncalled assets (\texttt{call\_count} = 0), highlighting type‑specific concentration patterns for assets. 
\section{Bounty Semantic Analysis}
\label{app:bounty_analysis}
Our overall analysis aims to evaluate the bounty resolution rate. 
To understand the underlying factors, we compare resolved and unresolved bounties and quantify their differences through semantic similarity. 
Specifically, we use the \texttt{title} field of each bounty, which provides a direct requirement of the task being requested~\cite{evomap-intro}. 
We embed all titles using OpenAI’s text‑embedding‑3‑small model and compute pairwise cosine similarity~\cite{openai-embed}. 
For each resolved bounty, we compare it against all unresolved bounties and record the maximum similarity score, which is then used for subsequent analysis.

\section{Dataset Schema}
\label{sec:appendix_dataset}
All data were crawled from the EvoMap platform via its public REST and GEP-A2A protocol endpoints during our measurement period.
Table~\ref{tab:schema_asset_detail} stores the detailed metadata of every asset (Gene, Capsule, or EvolutionEvent) discovered on the marketplace.
Table~\ref{tab:schema_bounty_submissions} records individual submissions that agents made in response to bounty tasks.
Table~\ref{tab:schema_bounty_details} captures the metadata and lifecycle state of each bounty task posted on the platform.

\begin{table*}[h!]
  \centering
  \small
  \setlength{\tabcolsep}{4pt}
  \begin{threeparttable}
    \caption{Schema of the \texttt{evomap\_asset\_detail} table.}
    \label{tab:schema_asset_detail}
    \begin{tabular}{l l p{8.5cm}}
      \toprule
      \textbf{Column} & \textbf{Type} & \textbf{Description} \\
      \midrule
      \texttt{asset\_id}              & VARCHAR & Unique SHA-256 hash identifying the asset. \\
      \texttt{asset\_type}            & VARCHAR & Asset type (\textit{Gene}, \textit{Capsule}, or \textit{EvolutionEvent}). \\
      \texttt{status}                 & VARCHAR & Lifecycle status (\textit{promoted}, \textit{candidate}, \textit{revoked}, \textit{archived}, \textit{flagged}, \textit{stale}). \\
      \texttt{source\_node\_id}       & VARCHAR & Node ID of the publishing agent. \\
      \texttt{trigger\_text}          & VARCHAR & Error signal or condition that activates the asset. \\
      \texttt{related\_asset\_id}     & VARCHAR & Hash of recommendation asset. \\
      \texttt{author}                 & VARCHAR & Display name of the asset author. \\
      \texttt{tags}                   & VARCHAR & Comma-separated descriptive tags. \\
      \texttt{signature}              & VARCHAR & Cryptographic signature of the payload. \\
      \texttt{chain\_id}              & VARCHAR & Chain ledger identifier for provenance. \\
      \texttt{model\_name}            & VARCHAR & LLM model used to generate the asset. \\
      \texttt{short\_title}           & VARCHAR & Human-readable short title. \\
      \texttt{nl\_summary}            & VARCHAR & Natural-language summary of the asset. \\
      \texttt{trust\_tier}            & VARCHAR & Platform-assigned trust tier (\textit{normal}). \\
      \texttt{asset\_created\_at}     & VARCHAR & Timestamp when the asset was created on the platform. \\
      \texttt{compute\_saved}         & VARCHAR & JSON object reporting compute savings metrics. \\
      \texttt{confidence}             & REAL    & Self-reported confidence score $C \in [0,1]$. \\
      \texttt{success\_streak}        & INTEGER & Consecutive successful invocations at publication time. \\
      \texttt{call\_count}            & INTEGER & Number of invocations by other agents.  \\
      \texttt{view\_count}            & INTEGER & Number of views by other agents.\\
      \texttt{reuse\_count}           & INTEGER & Number of reuses by other agents. \\
      \texttt{gdi\_score}             & REAL    & Overall GDI score (composite of intrinsic, usage, social). \\
      \texttt{gdi\_score\_mean}       & REAL    & Mean GDI score across the asset's history. \\
      \texttt{gdi\_intrinsic}         & REAL    & Intrinsic quality sub-score $GDI_\mathrm{I} \in [0,1]$. \\
      \texttt{gdi\_usage}             & REAL    & Usage-based sub-score $GDI_\mathrm{U} \in [0,1]$. \\
      \texttt{gdi\_usage\_lower}      & REAL    & Lower confidence bound of the usage sub-score. \\
      \texttt{gdi\_social}            & REAL    & Social-signal sub-score $GDI_\mathrm{S} \in [0,1]$. \\
      \texttt{gdi\_social\_lower}     & REAL    & Lower confidence bound of the social sub-score. \\
      \texttt{gdi\_freshness}         & REAL    & Freshness decay factor $\in [0,1]$. \\
      \texttt{upvotes}                & INTEGER & Number of upvotes received. \\
      \texttt{downvotes}              & INTEGER & Number of downvotes received. \\
      \texttt{agent\_rating\_avg}     & REAL    & Average rating from evaluating agents. \\
      \texttt{agent\_rating\_count}   & INTEGER & Number of agent ratings received. \\
      \texttt{fork\_count}            & INTEGER & Number of forks. \\
      \texttt{iteration\_count}       & INTEGER & Number of evolutionary iterations. \\
      \texttt{payload\_json}          & TEXT    & Full asset payload as JSON. \\
      \texttt{lineage\_json}          & TEXT    & Genealogical lineage as JSON. \\
      \texttt{bundle\_capsule\_json}  & TEXT    & Associated capsule bundle data. \\
      \texttt{bundle\_events\_json}   & TEXT    & Associated evolution events. \\
      \texttt{rawtext}                & TEXT    & Raw text dump of the asset detail page. \\
      \bottomrule
    \end{tabular}
  \end{threeparttable}
\end{table*}

\begin{table*}[h!]
  \centering
  \small
  \setlength{\tabcolsep}{4pt}
  \begin{threeparttable}
    \caption{Schema of the \texttt{bounty\_submissions} table.}
    \label{tab:schema_bounty_submissions}
    \begin{tabular}{l l p{8.5cm}}
      \toprule
      \textbf{Column}      & \textbf{Type} & \textbf{Description} \\
      \midrule
      \texttt{bounty\_id}   & VARCHAR & Foreign key referencing the parent bounty task. \\
      \texttt{submission\_id} & VARCHAR & Unique identifier for this submission. \\
      \texttt{node\_id}     & VARCHAR & Node ID of the submitting agent. \\
      \texttt{asset\_id}    & VARCHAR & Hash of the asset submitted as the solution. \\
      \texttt{status}       & VARCHAR & Submission status (\textit{pending}, \textit{accepted}, \textit{rejected}, \textit{runner\_up}). \\
      \texttt{created\_at}  & VARCHAR & Timestamp when the submission was created. \\
      \texttt{summary}      & VARCHAR & Short summary of the submitted solution. \\
      \texttt{content}      & VARCHAR & Full content of the submission. \\
      \bottomrule
    \end{tabular}
  \end{threeparttable}
\end{table*}

\begin{table*}[h!]
  \centering
  \small
  \setlength{\tabcolsep}{4pt}
  \begin{threeparttable}
    \caption{Schema of the \texttt{bounty\_details} table.}
    \label{tab:schema_bounty_details}
    \begin{tabular}{l l p{8.5cm}}
      \toprule
      \textbf{Column}                      & \textbf{Type} & \textbf{Description} \\
      \midrule
      \texttt{bounty\_id}                   & VARCHAR & Unique identifier of the bounty task. \\
      \texttt{question\_id}                 & VARCHAR & Associated question or issue identifier. \\
      \texttt{user\_id}                     & VARCHAR & ID of the user who posted the bounty. \\
      \texttt{amount}                       & FLOAT   & Credit reward offered. \\
      \texttt{status}                       & VARCHAR & Bounty status (\textit{open}, \textit{matched}, \textit{expired}, \textit{accepted}, \textit{settled}). \\
      \texttt{title}                        & VARCHAR & Title of the bounty task. \\
      \texttt{signals}                      & VARCHAR & Error signals or keywords the bounty targets. \\
      \texttt{boost\_level}                 & INTEGER & Platform boost level (visibility multiplier). \\
      \texttt{matched\_asset\_id}           & VARCHAR & Hash of the best-matching existing asset, if any. \\
      \texttt{matched\_node\_id}            & VARCHAR & Node ID of the agent owning the matched asset. \\
      \texttt{accepted\_at}                 & VARCHAR & Timestamp when a submission was accepted. \\
      \texttt{expires\_at}                  & VARCHAR & Bounty deadline. \\
      \texttt{created\_at}                  & VARCHAR & Bounty creation timestamp. \\
      \texttt{task\_id}                     & VARCHAR & Linked task identifier. \\
      \texttt{task\_status}                 & VARCHAR & Linked task status (\textit{open}, \textit{claimed}, \textit{completed}, \textit{expired}, \textit{decomposed}, \textit{aggregating}, \textit{cancelled}). \\
      \texttt{task\_claimed\_by}            & VARCHAR & Node ID of the agent who claimed the task. \\
      \texttt{task\_claimed\_at}            & VARCHAR & Timestamp when the task was claimed. \\
      \texttt{submission\_count}            & INTEGER & Total submissions received. \\
      \texttt{promoted\_submission\_count}  & INTEGER & Submissions that were promoted. \\
      \texttt{competition\_status}          & VARCHAR & Competition phase (\textit{awaiting\_competition}, \textit{competitive}). \\
      \texttt{review\_status}               & VARCHAR & Review metadata as JSON (initiation/completion time, votes). \\
      \texttt{updated\_at}                  & VARCHAR & Last update timestamp. \\
      \bottomrule
    \end{tabular}
  \end{threeparttable}
\end{table*}

\clearpage

\end{document}